\documentclass[11pt]{article}

\usepackage[preprint]{acl}

\usepackage{times}
\usepackage{latexsym}

\usepackage[T1]{fontenc}

\usepackage[utf8]{inputenc}

\usepackage{microtype}

\usepackage{inconsolata}

\usepackage{graphicx}
\usepackage{booktabs}
\usepackage{multirow}
\usepackage{amsmath}
\usepackage{subcaption}
\usepackage{amsfonts}

\title{CORTEX: Token-Level Hallucination Detection in RAG via Comparative Internal Representations}

\author{
Kazuaki Furumai\quad
Shuichiro Haruta\quad
Kazunori Matsumoto\quad
Daisuke Kamisaka\\
KDDI Research, Inc. \\
\texttt{\{ka-furumai, sh-haruta, da-kamisaka\}@kddi.com}
}

\begin{document}
\maketitle
\begin{abstract}

In this paper, we propose CORTEX, a token-level hallucination detection method for Retrieval-Augmented Generation (RAG). In long-form RAG outputs, hallucinations often arise in localized spans rather than throughout an entire response. CORTEX therefore identifies ungrounded content at the token level, enabling fine-grained localization of hallucinations.
The key intuition behind CORTEX is that tokens grounded in retrieved documents should be more strongly influenced by those documents than hallucinated tokens. To capture this document-induced effect, CORTEX compares internal representations of a large language model (LLM) under two conditions: with and without the retrieved documents. Instead of relying solely on each token's immediate sensitivity to the retrieved documents, CORTEX also leverages the propagation of document-grounded information through preceding tokens, reducing false positives for tokens whose evidence has already been absorbed into the context. Finally, CORTEX applies post-processing smoothing step that models the tendency of hallucination labels to persist over contiguous spans, reducing local noise and encouraging span-consistent predictions.
Experiments on two RAG benchmarks and three LLMs show that CORTEX substantially improves token-level hallucination detection, with each component consistently contributing to performance gains.

\end{abstract}

\section{Introduction}

\begin{figure*}[t]
\centering
  \includegraphics[width=\textwidth]{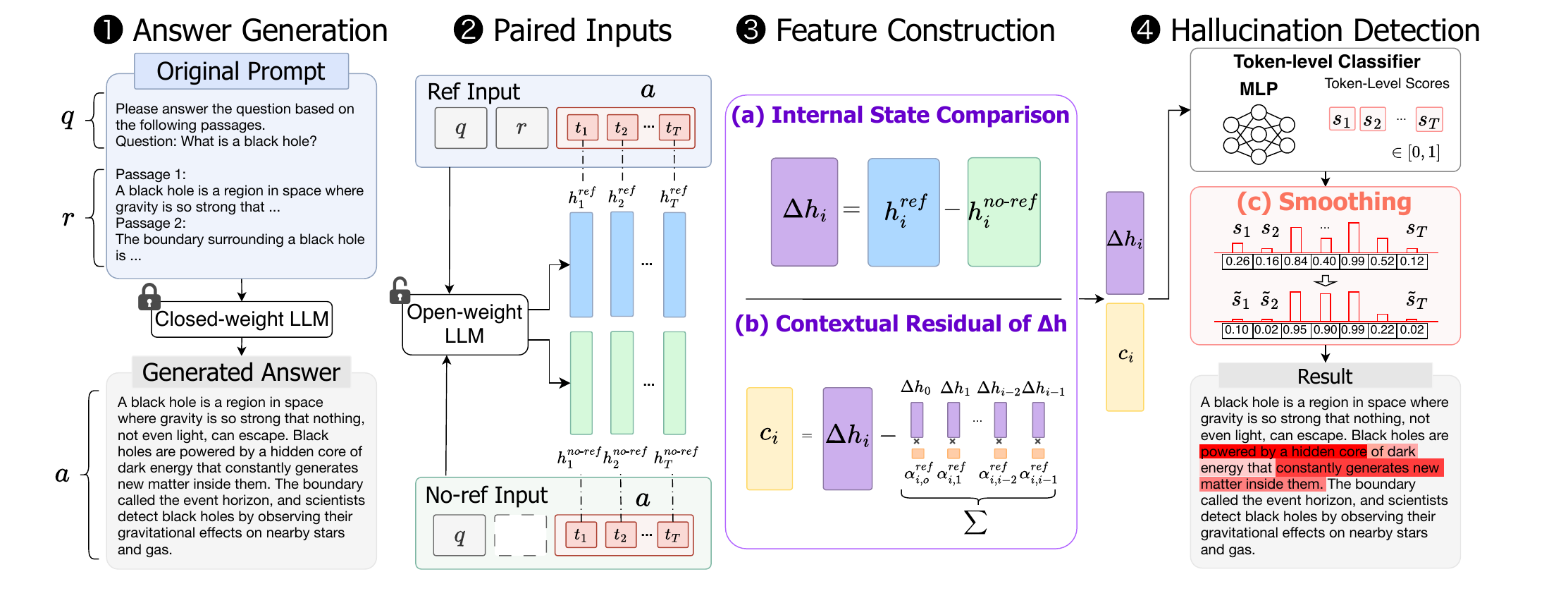}
  \caption{
Overview of CORTEX for token-level hallucination detection in RAG, using reference-conditioned internal representation comparisons, contextual residual features, and label-persistence smoothing.
} 
  \label{fig:overview}
\end{figure*}

Retrieval-Augmented Generation (RAG) has been widely adopted to improve the factuality of large language models (LLMs) by grounding generation in external references~\citep{NEURIPS2020_6b493230_rag}.
Although RAG mitigates hallucinations to some extent, LLMs may still generate groundless or inconsistent content even when references are provided~\citep{fan2026halluhardhardmultiturnhallucination}.
Accurately detecting such hallucinations within generated outputs therefore remains a critical challenge~\citep{niu-etal-2024-ragtruth, liu-etal-2025-attention,bang-etal-2025-hallulens}.

Various hallucination detection methods have been proposed to address this challenge.
Self-consistency-based approaches estimate reliability by measuring agreement across multiple outputs from the same prompt~\citep{manakul2023selfcheckgpt}.
Prompt-based methods use LLMs as fact-checkers to explicitly verify generated content~\citep{zheng2023llmasajudge, es-etal-2024-ragas, furumai-etal-2024-zero}.
While these approaches leverage LLMs' strong language understanding capabilities, they may incur high computational cost due to repeated sampling, and their reliability can depend on prompt design~\citep{ganesh-etal-2026-rethinking}.

More recently, hallucination detection methods based on the internal representations of LLMs have been increasingly studied~\citep{sriramanan2024llmcheck, zhang-etal-2025-icr, xiong2026toward}.
These methods seek hallucination-related signals in model activations and can offer insight into the internal conditions under which hallucinations arise.
This line of work has analyzed attention and feedforward-network states, quantified component contributions to generation, and extracted hallucination-related features from internal representations.

However, many existing approaches are not designed specifically for RAG and primarily assess the generated output as a whole, i.e., at the answer level.
This is limiting for long-form RAG outputs, where faithful and hallucinated content often coexist.
For practical use, detecting hallucinations in such mixed outputs requires a finer-grained formulation than answer-level detection.
The appropriate granularity can vary across tasks and annotation protocols, from words and phrases to sentences or paragraphs.
Given this variability, token-level detection can serve as a practical approach for localizing ungrounded content and later adapting predictions to the required granularity.

In this paper, we propose \textbf{CORTEX} (\textbf{C}omparative \textbf{O}bserved \textbf{R}eference-based \textbf{T}oken-level \textbf{EX}pressions), a post-hoc token-level hallucination detection method for RAG.
Figure~\ref{fig:overview} illustrates the overall framework.
CORTEX builds on the intuition that faithful tokens should exhibit coherent representational changes when references are provided, whereas hallucinated tokens should show weaker or less coherent changes.
To measure this effect, CORTEX constructs a paired counterfactual view of the same answer under reference-conditioned and no-reference inputs, rather than probing a single input as in conventional representation-based detectors.
The resulting reference-induced changes are then encoded as token-level delta features for hallucination detection.

CORTEX further introduces an attention-based contextual residual feature.
In long-form generation, reference influence may propagate indirectly through preceding answer tokens, as in reasoning-style or self-referential generation.
CORTEX captures this context-mediated reference influence by combining token-level delta features with attention patterns, helping the classifier distinguish tokens that are indirectly grounded in the reference from groundless tokens and thereby reducing false positives.

Finally, CORTEX uses post-processing smoothing step to adjust the granularity of token-level scores.
Raw token-level predictions can be sensitive to local noise and may produce isolated high-score tokens, whereas human hallucination annotations often appear as contiguous spans.
We therefore introduce \textit{label-persistence smoothing}, which treats raw scores as token-level confidence and models the tendency of hallucination labels to persist across neighboring tokens.
This reduces scattered noise and adapts token-level predictions to span-level annotation structure while still retaining token-level scores.

Experiments on two RAG benchmarks and three LLMs demonstrate that CORTEX substantially improves token-level hallucination detection, with all components contributing consistently to performance gains.

Our contributions are summarized as follows:

\begin{itemize}
    \item We propose CORTEX, which, to the best of our knowledge, is the first token-level hallucination detection method specifically designed for RAG.

    \item CORTEX is a practical post-hoc framework that is easy to implement and can detect hallucinations in outputs from arbitrary LLMs, including API-based closed-weight models.
\end{itemize}

\section{Preliminaries}

We consider a practical setting in which the answer is produced by a closed-weight LLM whose parameters and internal representations are not accessible, as is the case for many API-based models. 
Such models are often attractive in deployed applications because of their strong generation quality.
We denote this closed-weight LLM by $M_{\mathrm{close}}$. Given a question $q^{\mathrm{text}}$, the closed-weight LLM produces an answer as
\begin{equation}
a^{\mathrm{text}} = M_{\mathrm{close}}(q^{\mathrm{text}}).
\end{equation}

Our goal is to identify hallucinated tokens in the generated answer $a^{\mathrm{text}}$.
Although a direct way to obtain hallucination-related features would be to analyze the internal representations of the closed-weight LLM, we cannot do so for the aforementioned reason.
We therefore use a separate open-weight LLM, denoted by $M_{\mathrm{open}}$, as a post-hoc analysis model. 
Unlike $M_{\mathrm{close}}$, $M_{\mathrm{open}}$ exposes internal representations, which allows us to extract features from the answer in relation to the question.

Let $\mathrm{Tok}(\cdot)$ denote the tokenizer of $M_{\mathrm{open}}$. We tokenize the answer as
\begin{equation}
\begin{aligned}
a &= \mathrm{Tok}(a^{\mathrm{text}}) = (t_1, t_2, \ldots, t_i, \ldots, t_T),
\end{aligned}
\end{equation}
where $T$ denotes the number of answer tokens and $t_i$ denotes the $i$-th answer token under the tokenizer of $M_{\mathrm{open}}$.
We assume that $M_{\mathrm{open}}$ processes an input containing the question and the answer as $M_{\mathrm{open}}(q^{\mathrm{text}} \Vert a^{\mathrm{text}})$, where $\Vert$ denotes text concatenation with the appropriate prompt format.
$M_{\mathrm{open}}$ produces token-level internal representations over the entire input sequence, which capture the relationships between $q^{\mathrm{text}}$ and $a^{\mathrm{text}}$.
Since hallucination detection is performed over the answer, we use only the internal representations corresponding to the answer tokens. 
We denote the representation aligned with answer token $t_i$ by $h_i \in \mathbb{R}^d$. 
The specific input construction and the extraction of answer-token representations are described in Section~\ref{sec:cortex}.

The token-level hallucination detection task is to estimate whether each answer token is hallucinated. 
Let $y_i \in \{0,1\}$ denote the token-level label, where $y_i=1$ indicates that token $t_i$ is hallucinated and $y_i=0$ indicates that it is faithful. 

In our setting, we focus on RAG applications, where references are provided as grounding evidence for answers.
This setting reflects common practical deployments in which retrieved documents are used to reduce hallucination and the answer is expected to be faithful to those references.

\section{CORTEX}
\label{sec:cortex}
We propose CORTEX (\textbf{C}omparative \textbf{O}bserved \textbf{R}eference-based \textbf{T}oken-level \textbf{EX}pressions), a post-hoc hallucination detection framework for RAG.

CORTEX is built on three key ideas, illustrated in Figure~\ref{fig:overview}(a)--(c):
(a) a reference-induced delta representation for capturing how each answer token's internal representation changes when the references are provided;
(b) an attention-based contextual residual for distinguishing direct reference sensitivity from reference influence mediated by preceding answer tokens; 
and (c) label-persistence smoothing for reducing isolated token-level noise and obtaining span-consistent hallucination scores while preserving token-level predictions.

\subsection{Reference-Induced Delta Representation}

CORTEX constructs two conditioned inputs for the open-weight LLM, differing only in whether the references are included.
Let $r^{\mathrm{text}}$ denote the references, i.e., the retrieved documents used as grounding evidence in the RAG setting.
The reference-conditioned input is defined as $x_{\mathrm{ref}}^{\mathrm{text}} = q^{\mathrm{text}} \Vert r^{\mathrm{text}} \Vert a^{\mathrm{text}}$, whereas the no-reference input is defined as $x_{\mathrm{no\text{-}ref}}^{\mathrm{text}} = q^{\mathrm{text}} \Vert a^{\mathrm{text}}$.
With this construction, the answer span corresponds to the same token sequence $(t_1,\ldots,t_T)$ in both conditions, allowing CORTEX to compare internal representations aligned with the same answer tokens.

When $M_{\mathrm{open}}$ processes $x_{\mathrm{ref}}^{\mathrm{text}}$ and $x_{\mathrm{no\text{-}ref}}^{\mathrm{text}}$, it produces token-level internal representations over the entire input sequence. 
For an input $x = \mathrm{Tok}(x^{\mathrm{text}})$, let
$\mathrm{Rep}_{M_{\mathrm{open}}}(x)$
denote the sequence of internal representations obtained from $M_{\mathrm{open}}$.
We define an extraction function $f$ that selects the representations aligned with the answer tokens:
\begin{equation}
\begin{aligned}
(h^{\mathrm{ref}}_1, \ldots, h^{\mathrm{ref}}_T)
&=
f(\mathrm{Rep}_{M_{\mathrm{open}}}(x_{\mathrm{ref}})), \\
(h^{\mathrm{no\text{-}ref}}_1, \ldots, h^{\mathrm{no\text{-}ref}}_T)
&=
f(\mathrm{Rep}_{M_{\mathrm{open}}}(x_{\mathrm{no\text{-}ref}})).
\end{aligned}
\end{equation}

Here, $h^{\mathrm{ref}}_i, h^{\mathrm{no\text{-}ref}}_i \in \mathbb{R}^d$ denote the internal representations aligned with the same answer token $t_i$ under the reference-conditioned and no-reference conditions, respectively, and $d$ denotes the representation dimensionality.\footnote{Specifically, we use the final transformer layer output corresponding to each answer token as its token-level representation $h_i$.}

CORTEX builds on the intuition that faithful tokens should exhibit coherent representational changes when references are provided, whereas hallucinated tokens should show weaker or less coherent changes.
Based on this intuition, CORTEX computes the reference-induced delta representation for each answer token $t_i$ as
\begin{equation}
\Delta h_i
=
h_i^{\mathrm{ref}}
-
h_i^{\mathrm{no\text{-}ref}}.
\end{equation}

This vector is the core representation in CORTEX. Since the same answer is included in both inputs, $\Delta h_i$ captures how the representation of token $t_i$ changes when references are provided. 
In other words, $\Delta h_i$ is intended to emphasize the reference-induced change in how $M_{\mathrm{open}}$ contextualizes that token, rather than merely representing the token content itself.

\subsection{Attention-Based Contextual Residual}

The delta representation $\Delta h_i$ captures how the representation of token $t_i$ changes when references are added.
However, in reasoning-style outputs such as chain-of-thought, facts grounded in the references may first be stated in earlier parts of the answer, and later tokens may continue the reasoning based on those facts. 
In this case, the influence of the references can reach $t_i$ not only through the direct path $r \rightarrow t_i$, but also through the context-mediated path $r \rightarrow t_{<i} \rightarrow t_i$.
As a result, even tokens that are indirectly grounded by the references may appear to have a weak relationship with the references if we only use $\Delta h_i$.

To address this issue, we introduce a feature that represents how much reference influence is contained in the preceding tokens that the current token $t_i$ relies on as 
\begin{equation}
\bar{\Delta h}_i
=
\sum_{j<i}
\alpha_{ij}^{\mathrm{ref}} \Delta h_j ,
\end{equation}
where $\alpha_{ij}^{\mathrm{ref}} \in [0,1]$ denotes the attention weight from $t_i$ to a preceding answer token $t_j$ under the reference-conditioned input. 
We further subtract this context-mediated influence from the current token's own change and define the contextual residual as
\begin{equation}
c_i
=
\Delta h_i
-
\bar{\Delta h}_i .
\end{equation}
This residual represents the token-specific change that remains after removing the reference influence explainable through preceding tokens. We use both $\Delta h_i$ and $c_i$ for token-level hallucination detection.

\subsection{Token-Level Hallucination Detection}

For each answer token $t_i$, CORTEX predicts a raw token-level hallucination score $s_i$ by feeding $[\Delta h_i;c_i]$ into a multilayer perceptron (MLP) classifier $g_\theta$:
\begin{equation}
s_i
=
\sigma\!\left(g_\theta\!\left([\Delta h_i ; c_i]\right)\right),
\qquad
s_i \in [0,1],
\end{equation}
where $\sigma$ denotes the sigmoid function.
The classifier is trained with binary cross-entropy loss using token-level hallucination labels.
To address class imbalance, the positive class is weighted according to the ratio of negative to positive tokens in the training set.

\subsection{Label-Persistence Smoothing}
\label{sec:label_persistence_smoothing}
We define a span as a contiguous sequence of tokens annotated as a single hallucination unit.
Human hallucination annotations are typically span-based: once a token is marked as hallucinated, neighboring tokens in the same span are likely to receive the same label.
Based on this property, we apply \textit{label-persistence smoothing} as a post-processing step, rather than using moving-average smoothing.

Given the raw token-level scores $s_{1:T}$, we introduce an unobserved binary smoothing-label variable $z_i\in\{0,1\}$ to represent the span-consistent label underlying the post-processed score, where $z_i=0$ denotes a faithful token and $z_i=1$ denotes a hallucinated token.
To obtain span-consistent post-processed scores, we model the smoothed label sequence by combining two components around positions $i$ and $i+1$: the tendency of neighboring labels to persist and information from the raw scores at each position.

We first model span-level label persistence by defining the pairwise persistence term between neighboring labels as
\begin{equation}
\label{eq:persistence}
\rho(z_i,z_{i+1})
=
\begin{cases}
p_{\mathrm{stay}}, & z_{i+1}=z_i,\\
1-p_{\mathrm{stay}}, & z_{i+1}\neq z_i.
\end{cases}
\end{equation}
The parameter $p_{\mathrm{stay}}\in[0,1]$ controls the degree of label persistence: smaller values allow finer-grained label changes, whereas larger values favor longer same-label spans.
This provides a deliberately simple approximation to span-level label continuity, rather than modeling the full variability of human annotation patterns. 
We then define the token-level confidence induced by the raw score $s_i$ at each position:
\begin{equation}
\label{eq:token_confidence}
\phi_i(z_i=1)=s_i,
\qquad
\phi_i(z_i=0)=1-s_i.
\end{equation}
This quantity represents how the raw score at position $i$ is converted into token-level confidence for each value of the smoothing-label variable.\footnote{In implementation, we apply a small amount of clipping to the raw scores so that the token-level confidence does not become exactly 0 or 1.}

Combining Eq.~\eqref{eq:persistence} and Eq.~\eqref{eq:token_confidence}, we define the local compatibility term for neighboring positions $i$ and $i+1$ as
\begin{equation}
\label{eq:positions_i_i_plus_1}
\phi_i(z_i)\rho(z_i,z_{i+1})\phi_{i+1}(z_{i+1}).
\end{equation}
This term measures the compatibility of the neighboring label assignment $(z_i,z_{i+1})$ with both the raw scores and the label-persistence assumption.

We therefore define the normalized distribution over sequence labels as
\begin{equation}
\label{eq:sequence_label_distribution}
\begin{aligned}
P(z_{1:T}\mid s_{1:T})
&=
\frac{1}{Z(s_{1:T})}
\pi(z_1)
\prod_{i=1}^{T}
\phi_i(z_i)
\\
&\quad\times
\prod_{i=1}^{T-1}
\rho(z_i,z_{i+1}),
\end{aligned}
\end{equation}
where $Z(s_{1:T})$ is the normalizing constant and $\pi(z_1)$ is the initial distribution over the first smoothing label.
We use a uniform initial distribution, $\pi(0)=\pi(1)=1/2$.

Although $z_{1:T}$ is unobserved, the desired token-level smoothed score can be obtained by marginalizing over all smoothing-label sequences:
\begin{equation}
\label{eq:smoothed_score}
\tilde{s}_i=P(z_i=1\mid s_{1:T}).
\end{equation}

This posterior can be computed efficiently using the forward--backward algorithm~\citep{rabiner1989tutorial}.
We provide the full recursions and posterior marginal formula in Appendix~\ref{app:label_persistence_smoothing}.

\section{Experiments}

We evaluate CORTEX on two publicly available RAG hallucination benchmarks, RAGTruth~\citep{niu-etal-2024-ragtruth} and HalluRAG~\citep{ridder2024halluragdatasetdetectingcloseddomain}, using three LLMs: Llama-3.1-8B-Instruct~\citep{grattafiori2024llama3herdmodels}, Qwen3-8B~\citep{yang2025qwen3technicalreport}, and Mistral-7B-Instruct-v0.2~\citep{mistralai2023mistral7binstructv02}.
We use these open-weight LLMs to obtain internal representations for CORTEX and the relevant baselines.
Evaluation is performed at the token and answer levels.
Token-level evaluation assesses whether a method can identify hallucinated tokens in a generated answer, while answer-level evaluation assesses whether the answer contains any hallucinated content.
For CORTEX, answer-level evaluation is performed by aggregating token-level scores, as it is trained for token-level detection and does not use answer-level supervision.
Specifically, we use the maximum token-level hallucination score as the answer-level score.

For label-persistence smoothing, we use $p_{\mathrm{stay}}=0.993$ at the token level and $p_{\mathrm{stay}}=0.930$ at the answer level.
We provide a sensitivity analysis of $p_{\mathrm{stay}}$ in Appendix~\ref{sec:pstay_sensitivity}.
Additional details on the experimental setup are provided in Appendix~\ref{sec:experimental_details}.
The following subsections describe the baselines and datasets used in our experiments.

\subsection{Baselines}

We compare CORTEX with five baselines.
\textbf{NLL} uses negative log-likelihood as an uncertainty-based signal, following prior work on uncertainty estimation from next-token predictive distributions~\citep{Malinin2021UncertaintyEI}.
\textbf{SAPLMA} trains a probe on intermediate transformer representations, based on the observation that hidden states encode hallucination-related signals~\citep{azaria-mitchell-2023-internal}.
\textbf{LLM-Check} extracts features from internal states produced by attention mechanisms and feedforward neural networks, including spectral properties such as eigenvalues~\citep{sriramanan2024llmcheck}.
\textbf{ICR Probe} uses internal component attribution to quantify the contribution of attention and feedforward networks and identify hallucinated content~\citep{zhang-etal-2025-icr}.
\textbf{RAGLens} applies sparse autoencoders to token embeddings and selects hallucination-related sparse features via token aggregation and mutual-information-based feature selection~\citep{xiong2026toward}.

For answer-level evaluation, we follow the original implementation of baselines. However, their answer-level implementation cannot be directly used for token-level evaluation. We modify parts of the baseline implementations to adapt them to token-level detection while preserving their original mechanisms as much as possible.

\subsection{Datasets}

Datasets for hallucination detection in RAG settings remain limited, especially those that include both references and fine-grained hallucination annotations.
To the best of our knowledge, RAGTruth is the only publicly available RAG benchmark with human-annotated hallucination spans that are sufficiently fine-grained to support token-level evaluation.
Although RAGTruth contains multiple task types, we use only its QA subset in this work.

HalluRAG consists of answers generated by multiple LLMs under two RAG prompt settings using either relevant or irrelevant Wikipedia chunks, with sentence-level hallucination annotations by GPT-4o~\citep{openai2024gpt4ocard}.
Since token-level labels are unavailable, we derive token-level pseudo-labels by marking all tokens in hallucinated sentences as hallucinated.

The two datasets provide complementary settings, differing in annotation granularity, labeling procedure, retrieval quality, and QA distribution.
RAGTruth provides fine-grained human span annotations, whereas HalluRAG provides sentence-level GPT-4o annotations and includes cases where references may be irrelevant to the question.
They also differ in construction and topical coverage, with RAGTruth's QA subset is based on daily-life questions, whereas HalluRAG derives questions from Wikipedia passages.
Together, these differences allow us to assess robustness across conditions.

Both datasets contain answers generated by various LLMs, together with the references provided for generating those answers.
We treat these answers as outputs from closed-weight LLMs, regardless of which model generated them.
To ensure a fair comparison, all methods receive the same answer and corresponding references as input.

\begin{table}[t]
\centering
\small
\begin{tabular}{lcc}
\hline
Statistic & RAGTruth & HalluRAG \\
\hline
\# Samples & 5{,}934 & 2{,}243 \\
Train samples & 4{,}530 & 1{,}662 \\
Validation samples & 504 & 185 \\
Test samples & 900 & 396 \\
Avg. prompt len. & 1{,}660.0 & 1{,}778.9 \\
Avg. answer len. & 686.5 & 164.0 \\
\hline
\end{tabular}
\caption{Dataset statistics.}
\label{tab:dataset_stats}
\end{table}

\subsection{Results}
\subsection*{Token-level Detection}

\begin{table*}[t]
\centering
\small
\setlength{\tabcolsep}{3pt}
\renewcommand{\arraystretch}{1.1}
\begin{tabular}{l|cc|cc|cc|cc|cc|cc}
\toprule
\multirow{3}{*}{\textbf{Method}} 
& \multicolumn{6}{c|}{\textbf{RAGTruth}} 
& \multicolumn{6}{c}{\textbf{HalluRAG}} \\
\cmidrule(lr){2-7} \cmidrule(lr){8-13}

& \multicolumn{2}{c|}{\textbf{Llama}} 
& \multicolumn{2}{c|}{\textbf{Qwen}} 
& \multicolumn{2}{c|}{\textbf{Mistral}} 
& \multicolumn{2}{c|}{\textbf{Llama}} 
& \multicolumn{2}{c|}{\textbf{Qwen}} 
& \multicolumn{2}{c}{\textbf{Mistral}} \\
\cmidrule(lr){2-3} \cmidrule(lr){4-5} \cmidrule(lr){6-7}
\cmidrule(lr){8-9} \cmidrule(lr){10-11} \cmidrule(lr){12-13}

& AP & AUROC
& AP & AUROC
& AP & AUROC
& AP & AUROC
& AP & AUROC
& AP & AUROC \\
\midrule

NLL
& 0.0476 & 0.4676 & 0.0495 & 0.4686 & 0.0479 & 0.4666 
& 0.2648 & 0.4952 & 0.2658 & 0.4934 & 0.2685 & 0.4945 \\

LLM-Check
& 0.0769 & 0.6079 & 0.0951 & 0.6480 & 0.0743 & 0.6189
& 0.2994 & 0.5411 & 0.2874 & 0.5342 & 0.2944 & 0.5467 \\

ICR Probe
& 0.2665 & 0.5380 & 0.2464 & 0.5067 & 0.2621 & 0.5311 
& 0.3292 & 0.5285 & 0.3131 & 0.5106 & 0.3197 & 0.5143 \\

SAPLMA
& 0.3894 & 0.8879 & 0.4182 & 0.8820 & 0.4027 & 0.8822
& \underline{0.6875} & \underline{0.8024} & \underline{0.6584} & \underline{0.8129} & \underline{0.6193} & \underline{0.7578} \\

RAGLens
& \underline{0.4271} & \underline{0.8953} & \underline{0.4575} & \underline{0.9075} & \underline{0.4244} & \underline{0.8884}
& 0.5725 & 0.7425 & 0.6380 & 0.7639 & 0.5596 & 0.7393 \\

\textbf{CORTEX}

& \textbf{0.5686} & \textbf{0.9275} & \textbf{0.5940} & \textbf{0.9372} & \textbf{0.5473} & \textbf{0.9244} 
& \textbf{0.7690} & \textbf{0.8360} & \textbf{0.7495} & \textbf{0.8426} & \textbf{0.7308} & \textbf{0.8233} \\

\bottomrule
\end{tabular}

\caption{Token-level hallucination detection results on RAGTruth and HalluRAG. Bold and underline indicate the best and second-best results, respectively.}
\label{tab:main_results_token_level}

\end{table*}

\begin{table*}[th]
\centering
\small
\setlength{\tabcolsep}{3pt}
\renewcommand{\arraystretch}{1.1}

\begin{tabular}{l|cc|cc|cc|cc|cc|cc}
\toprule
\multirow{3}{*}{\textbf{Method}} 
& \multicolumn{6}{c|}{\textbf{RAGTruth}} 
& \multicolumn{6}{c}{\textbf{HalluRAG}} \\
\cmidrule(lr){2-7} \cmidrule(lr){8-13}

& \multicolumn{2}{c|}{\textbf{Llama}} 
& \multicolumn{2}{c|}{\textbf{Qwen}} 
& \multicolumn{2}{c|}{\textbf{Mistral}} 
& \multicolumn{2}{c|}{\textbf{Llama}} 
& \multicolumn{2}{c|}{\textbf{Qwen}} 
& \multicolumn{2}{c}{\textbf{Mistral}} \\
\cmidrule(lr){2-3} \cmidrule(lr){4-5} \cmidrule(lr){6-7}
\cmidrule(lr){8-9} \cmidrule(lr){10-11} \cmidrule(lr){12-13}

& AP & AUROC 
& AP & AUROC 
& AP & AUROC 
& AP & AUROC  
& AP & AUROC 
& AP & AUROC \\
\midrule

NLL
& 0.1994 & 0.5621 & 0.2019 & 0.5227 & 0.2313 & 0.6088 
& 0.2055 & 0.4827 & 0.3184 & 0.5806 & 0.2969& 0.5600 \\

LLM-Check
& 0.3375 & 0.7100 & 0.3330 & 0.7198 & 0.3240 & 0.7170
& 0.3052 & 0.5711 & 0.3956 & 0.6638 & 0.4059 & 0.6481 \\

ICR Probe
& 0.1286 & 0.3459 & 0.2301 & 0.6162 & 0.1831 & 0.5182
& 0.4077 & 0.6640 & 0.3286 & 0.6031 & 0.2157 & 0.4310 \\

SAPLMA
& 0.6708 & 0.8863 & 0.5901 & 0.8600 & 0.4897 & 0.8077
& \underline{0.7922} & \underline{0.8482} & \underline{0.7867} & \textbf{0.8695} & 0.7529 & 0.8628 \\

RAGLens
& \textbf{0.7329} & \textbf{0.9011} & \underline{0.6679} & \underline{0.8812} & \underline{0.6781} & \underline{0.8819}
& \textbf{0.7994} & \textbf{0.8905} & 0.7669 & 0.8380 & \underline{0.8006} & \textbf{0.8835} \\

\textbf{CORTEX}


& \underline{0.6893} & \underline{0.8957} & \textbf{0.7596} & \textbf{0.9077} & \textbf{0.6787} & \textbf{0.8907}
& 0.7795 & 0.8481 & \textbf{0.7907} & \underline{0.8627} & \textbf{0.8010} & \underline{0.8718} \\

\bottomrule
\end{tabular}

\caption{
Answer-level hallucination detection results. Bold and underline indicate the best and second-best results, respectively.
}
\label{tab:main_results_answer_level}

\end{table*}
Table~\ref{tab:main_results_token_level} reports token-level hallucination detection results in terms of average precision (AP) and the area under the receiver operating characteristic curve (AUROC).
Across both RAGTruth and HalluRAG, CORTEX achieves the best performance in all settings.
The gains are particularly substantial on RAGTruth, where human fine-grained annotations are available, suggesting that CORTEX is well aligned with token-level hallucination detection.
Even on HalluRAG, where token-level labels are derived from sentence-level annotations, CORTEX remains the best-performing method, indicating robustness to coarser and noisier supervision.

All baselines receive inputs with references, but rely on a single reference-conditioned view of model representations.
SAPLMA directly uses transformer layer outputs as features, while RAGLens extracts hallucination-related sparse features from token embeddings using a trained sparse encoder.
In contrast, CORTEX derives its signal from the contrast between paired representations of the same tokens with and without references.
The consistent gains suggest that this comparative formulation captures reference-induced hallucination signals that are not readily available from a single internal representation or features derived from it.

Label-persistence smoothing is a post-processing module applicable to any token-level predictions.
In Appendix~\ref{sec:appendix_smoothing_baselines}, we further apply it to baselines to demonstrate its general effectiveness.

\subsection*{Answer-level Detection}
Table~\ref{tab:main_results_answer_level} reports the answer-level hallucination detection results in terms of AP and AUROC.
Unlike the baselines, which are trained directly for answer-level detection, CORTEX does not use answer-level supervision.
Instead, CORTEX reuses the token-level hallucination scores and assigns each answer the maximum score among its tokens.
Thus, this setting evaluates whether the token-level signals captured by CORTEX can also indicate hallucination at the answer level.

Despite this simple aggregation strategy, CORTEX achieves competitive performance against the baselines.
This suggests that its token-level predictions provide useful signals for both localized detection and answer-level reliability assessment.
At the same time, the strong performance of some answer-level baselines suggests that answer-level detection may require global signals beyond localized token-level signals.
Taken together, these results indicate that token-level and answer-level hallucination detection are both important and should be studied as complementary problems.

\subsection{Ablation Study}

\begin{table}[t]
\centering
\small
\setlength{\tabcolsep}{3pt}
\begin{tabular}{lcccc}
\hline
\multirow{2}{*}{Method}
& \multicolumn{2}{c}{Token-level}
& \multicolumn{2}{c}{Answer-level} \\
\cline{2-5}
& AP & AUROC & AP & AUROC \\
\hline
$\Delta h$ 
& 0.4732 & 0.9074 & 0.6040 & 0.8853 \\
$\Delta h+c$
& 0.5335 & 0.9156 & 0.6910 & 0.9011 \\
$\Delta h+c+\mathrm{Smooth}$
& 0.5940 & 0.9372 & 0.7596 & 0.9077 \\
\hline
\end{tabular}
\caption{Ablation results on RAGTruth using Qwen.}
\label{tab:ablation_ragtruth_qwen}
\end{table}

\begin{figure*}[t]
  \includegraphics[width=\textwidth]{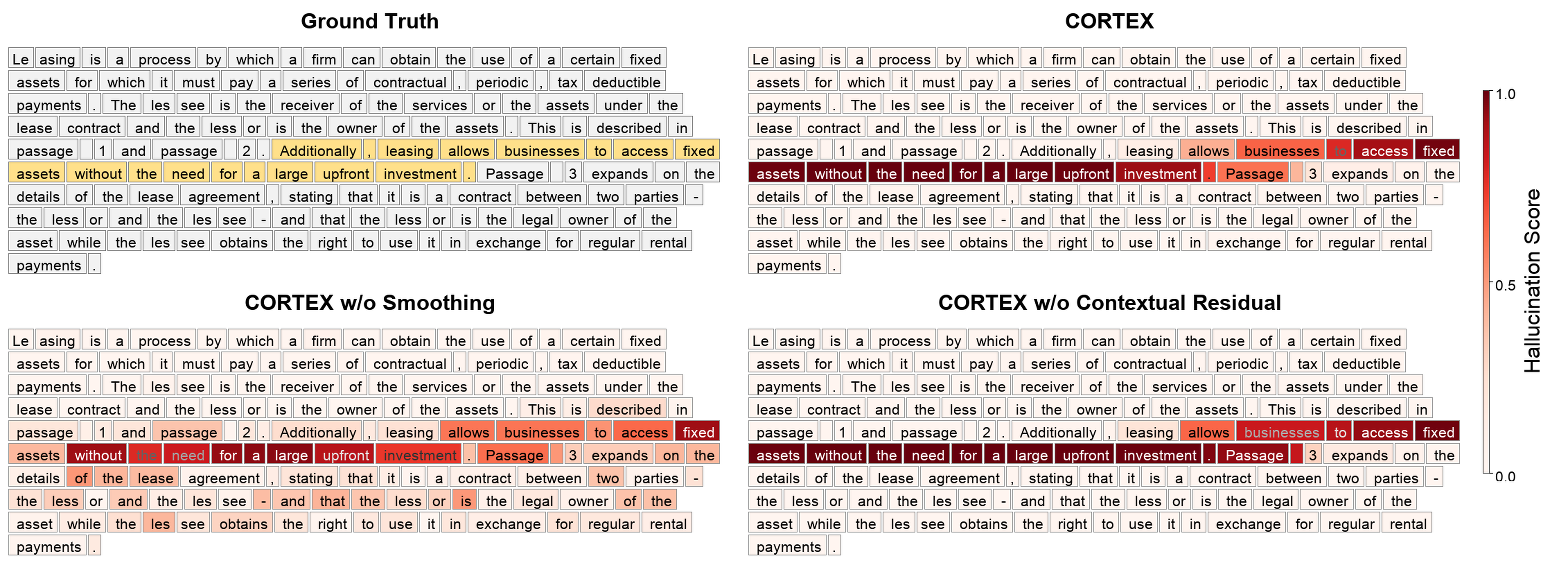}
  \caption{Detection example for an answer containing hallucinated content. Label-persistence smoothing suppresses noisy token-level score patterns and highlights the hallucinated span more accurately and coherently.}
  \label{fig:ablation_heatmap_smoothing}
\end{figure*}

\begin{figure*}[t]
  \includegraphics[width=\textwidth]{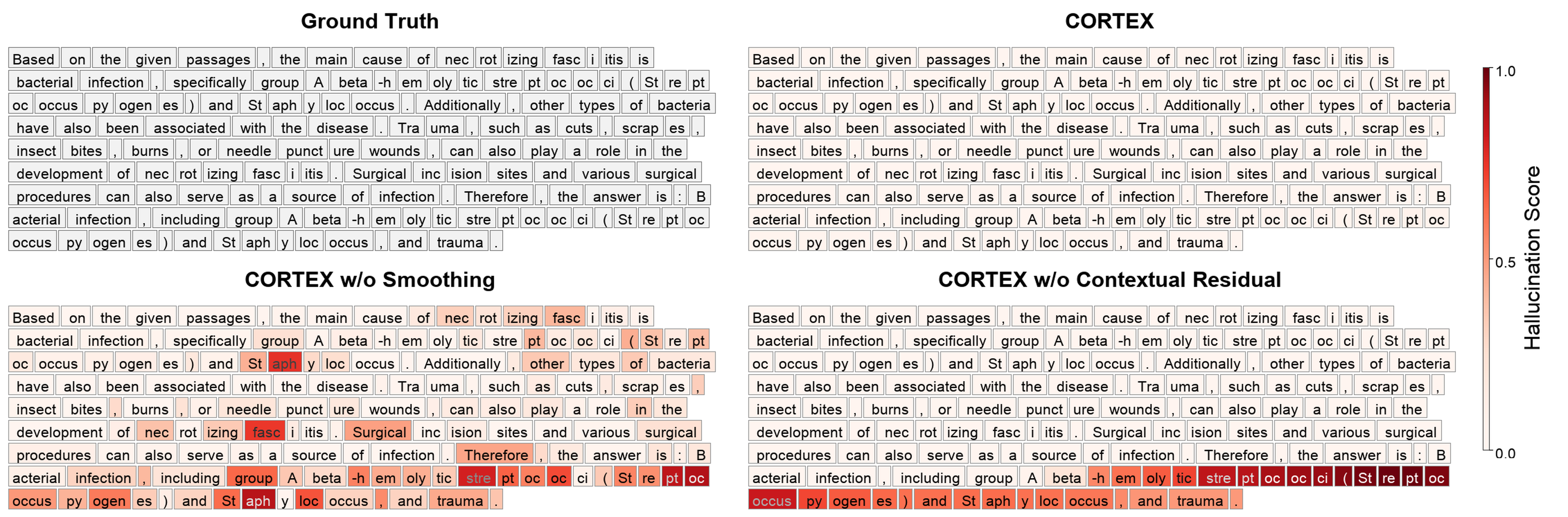}
  \caption{Detection example for an answer without hallucinated content. The contextual residual $c$ reduces false positives by accounting for reference influence mediated through the preceding answer context.}
  \label{fig:ablation_heatmap_ci}
\end{figure*}

To analyze the contribution of each CORTEX component, we conduct an ablation study.
Table~\ref{tab:ablation_ragtruth_qwen} reports token- and answer-level performance for each configuration.
Both the contextual residual $c$ and label-persistence smoothing ($\mathrm{Smooth}$) improve performance.
Removing label-persistence smoothing shows that raw token-level scores are informative but locally unstable, while removing $c$ shows that context-mediated reference influence provides information complementary to $\Delta h$.

Figures~\ref{fig:ablation_heatmap_smoothing} and~\ref{fig:ablation_heatmap_ci} illustrate these effects qualitatively using cases with and without hallucinations.
In the heatmaps, color intensity reflects the predicted hallucination score, with redder tokens indicating higher hallucination likelihood.
In Figure~\ref{fig:ablation_heatmap_smoothing}, the model without label-persistence smoothing produces interleaved high- and low-score tokens across semantically coherent regions, yielding predictions less consistent with human span-level annotations.
In contrast, full CORTEX produces smoother and more contiguous high-score regions, better matching the ground-truth hallucination spans.
Figure~\ref{fig:ablation_heatmap_ci} illustrates the role of the contextual residual $c$.
Without $c$, the model can overemphasize the absence of direct reference influence and assign high scores to tokens indirectly grounded through the preceding answer context.
With $c$, CORTEX accounts for reference influence propagated through previous tokens and reduces false positives caused by indirect grounding.

Overall, these ablation results support the design of CORTEX.
The combination of $\Delta h$, the contextual residual $c$, and label-persistence smoothing enables CORTEX to achieve robust localized hallucination detection.
Further ablation case studies are presented in Appendix~\ref{sec:appendix_ablation_case_studies}.

\section{Related Work}

The problem of hallucination in LLMs has been extensively studied, yet remains unresolved~\citep{Huang2025survey, kalai2025languagemodelshallucinate}.
It undermines the reliability of LLM-based AI agents and hinders real-world deployment.
RAG mitigates hallucinations by incorporating references into prompts~\citep{gao2024retrievalaugmentedgenerationlargelanguage, Rackauckas_2024}, but LLMs may still generate hallucinated content even with references, motivating the detection of groundless or inconsistent RAG outputs~\citep{fan2026halluhardhardmultiturnhallucination}.

Many hallucination detection methods rely on external verification or repeated generation.
Self-consistency methods estimate reliability by comparing multiple outputs from the same prompt~\citep{manakul2023selfcheckgpt}, while prompt-based approaches use LLMs as judges or fact-checkers~\citep{li-etal-2025-generation, es-etal-2024-ragas, furumai-etal-2024-zero}.
These methods exploit LLMs' language understanding but depend on prompt design and verifier capability, and often require additional inference.

Another line of work uses model-internal signals, including uncertainty from next-token distributions~\citep{Malinin2021UncertaintyEI}, hallucination-related information in intermediate transformer representations~\citep{azaria-mitchell-2023-internal}, spectral features of attention and feedforward states~\citep{sriramanan2024llmcheck}, component-level attribution for tracing information sources~\citep{zhang-etal-2025-icr}, and sparse autoencoder features for RAG hallucination analysis~\citep{xiong2026toward}.
These studies show that hallucination-related signals are present in model computations and representations, motivating representation-based detection.

CORTEX differs from prior representation-based approaches by comparing internal representations obtained from reference-conditioned and no-reference inputs, rather than analyzing a single internal state in isolation.
It further uses label-persistence smoothing to reduce isolated noise and better align predictions with span-based hallucination annotations.

\section{Conclusion}

We proposed CORTEX, a post-hoc token-level hallucination detection method for RAG.
CORTEX constructs a paired counterfactual view of the same answer by analyzing its internal representations with and without references, and encodes reference-induced changes as token-level delta features.
It combines these features with attention patterns to capture context-mediated reference influence and uses label-persistence smoothing to reduce local noise while preserving span-consistent scores.

Experiments show that CORTEX substantially outperforms token-level baselines and achieves competitive answer-level performance through simple score aggregation.
Ablations confirm the contributions of both the attention-based contextual residual and label-persistence smoothing.
These results demonstrate the effectiveness of comparing internal representations for hallucination detection in RAG.
By using an open-weight LLM to analyze outputs from closed-weight LLMs without modifying generation, CORTEX provides a practical approach to reliability assessment.
Beyond detection, these comparative internal-representation signals may provide a basis for future hallucination mitigation, including reward modeling for tuning generators and objectives that encourage stronger reference-induced representations.

\section*{Limitations}

CORTEX is designed for RAG settings and therefore assumes the presence of references against which generated outputs can be assessed. It is not intended to verify claims that are generated solely from the parametric knowledge of an LLM.
This restriction may be acceptable, or even desirable, in controlled applications such as enterprise customer support, where answers should be grounded only in approved documents.
In more open-ended assistant scenarios, however, this may be limiting because useful information that is not explicitly grounded in the provided references may be treated as groundless.

Another limitation is that CORTEX requires access to an open-weight LLM that exposes internal representations, including hidden states and attention weights.
Although the original generator can be API-based or otherwise inaccessible, the post-hoc analysis model must provide sufficient internal signals.
The quality of detection may therefore depend on the choice of the open-weight LLM and its ability to interpret the generated answer and references.

A further limitation is the need for fine-grained supervision during training.
Although RAGTruth provides human span-level annotations, such annotations remain scarce for RAG hallucination detection.
Improving supervision under limited annotation resources remains an important direction for future work.

\section*{Ethical Considerations}
CORTEX is intended to support reliability assessment in RAG systems by identifying potentially groundless tokens in generated outputs.
Such detection can help reduce the risk of users relying on hallucinated information, especially in applications where answers are expected to be grounded in specific references.
However, CORTEX should not be interpreted as a guarantee of factual correctness.
Its predictions are probabilistic and may include both false positives and false negatives; therefore, human review or additional verification may still be necessary in high-stakes domains.

There is also a risk that hallucination detection tools could be over-relied upon or used to present generated outputs as more reliable than they actually are.
We encourage practitioners to communicate the limitations of detection results clearly and to avoid using CORTEX as the sole safety mechanism for critical decision-making.

It is also important to distinguish the scope of CORTEX from broader safety evaluation.
CORTEX is designed to detect ungrounded content in reference-grounded generation, not to determine broader notions of truth, fairness, or social harm.
Future work should examine how token-level reliability signals can be combined with other safeguards, including checks for bias, toxicity, privacy risks, and domain-specific safety issues.
\bibliography{custom}

@inproceedings{manakul2023selfcheckgpt,
    title = {{S}elf{C}heck{GPT}: Zero-Resource Black-Box Hallucination Detection for Generative Large Language Models},
    author = {Manakul, Potsawee  and
      Liusie, Adian  and
      Gales, Mark},
    booktitle = {Proceedings of the 2023 Conference on Empirical Methods in Natural Language Processing},
    year = {2023},
    publisher = {Association for Computational Linguistics},
    url = {https://aclanthology.org/2023.emnlp-main.557/},
    doi = {10.18653/v1/2023.emnlp-main.557},
    pages = {9004--9017},
}

@inproceedings{li-etal-2025-generation,
    title = "From Generation to Judgment: Opportunities and Challenges of {LLM}-as-a-judge",
    author = "Li, Dawei  and
      Jiang, Bohan  and
      Huang, Liangjie  and
      Beigi, Alimohammad  and
      Zhao, Chengshuai  and
      Tan, Zhen  and
      Bhattacharjee, Amrita  and
      Jiang, Yuxuan  and
      Chen, Canyu  and
      Wu, Tianhao  and
      Shu, Kai  and
      Cheng, Lu  and
      Liu, Huan",
    editor = "Christodoulopoulos, Christos  and
      Chakraborty, Tanmoy  and
      Rose, Carolyn  and
      Peng, Violet",
    booktitle = "Proceedings of the 2025 Conference on Empirical Methods in Natural Language Processing",
    year = "2025",
    publisher = "Association for Computational Linguistics",
    url = "https://aclanthology.org/2025.emnlp-main.138/",
    doi = "10.18653/v1/2025.emnlp-main.138",
    pages = "2757--2791",
    ISBN = "979-8-89176-332-6"
}

@inproceedings{zheng2023llmasajudge,
    author = {Zheng, Lianmin and Chiang, Wei-Lin and Sheng, Ying and Zhuang, Siyuan and Wu, Zhanghao and Zhuang, Yonghao and Lin, Zi and Li, Zhuohan and Li, Dacheng and Xing, Eric P. and Zhang, Hao and Gonzalez, Joseph E. and Stoica, Ion},
    title = {Judging LLM-as-a-judge with MT-bench and Chatbot Arena},
    year = {2023},
    url = {https://proceedings.neurips.cc/paper_files/paper/2023/file/91f18a1287b398d378ef22505bf41832-Paper-Datasets_and_Benchmarks.pdf},
    publisher = {Curran Associates Inc.},
    booktitle = {Proceedings of the 37th International Conference on Neural Information Processing Systems},
    articleno = {2020},
    numpages = {29},
    }

@inproceedings{es-etal-2024-ragas,
    title = "{RAGA}s: Automated Evaluation of Retrieval Augmented Generation",
    author = "Es, Shahul  and
      James, Jithin  and
      Espinosa Anke, Luis  and
      Schockaert, Steven",
    editor = "Aletras, Nikolaos  and
      De Clercq, Orphee",
    booktitle = "Proceedings of the 18th Conference of the European Chapter of the Association for Computational Linguistics: System Demonstrations",
    year = "2024",
    publisher = "Association for Computational Linguistics",
    url = "https://aclanthology.org/2024.eacl-demo.16/",
    doi = "10.18653/v1/2024.eacl-demo.16",
    pages = "150--158",
}

@inproceedings{sriramanan2024llmcheck,
author = {Sriramanan, Gaurang and Bharti, Siddhant and Sadasivan, Vinu Sankar and Saha, Shoumik and Kattakinda, Priyatham and Feizi, Soheil},
title = {LLM-check: investigating detection of hallucinations in large language models},
year = {2024},
isbn = {9798331314385},
url = {https://proceedings.neurips.cc/paper_files/paper/2024/file/3c1e1fdf305195cd620c118aaa9717ad-Paper-Conference.pdf},
publisher = {Curran Associates Inc.},
booktitle = {Proceedings of the 38th International Conference on Neural Information Processing Systems},
articleno = {1077},
numpages = {29},
}

@inproceedings{zhang-etal-2025-icr,
    title = {{ICR} Probe: Tracking Hidden State Dynamics for Reliable Hallucination Detection in {LLM}s},
    author = {Zhang, Zhenliang  and
      Hu, Xinyu  and
      Zhang, Huixuan  and
      Zhang, Junzhe  and
      Wan, Xiaojun},
    booktitle = {Proceedings of the 63rd Annual Meeting of the Association for Computational Linguistics},
    year = {2025},
    url = {https://aclanthology.org/2025.acl-long.880/},
    publisher = {Association for Computational Linguistics},
    pages = {17986--18002},
}

@article{Huang2025survey,
author = {Huang, Lei and Yu, Weijiang and Ma, Weitao and Zhong, Weihong and Feng, Zhangyin and Wang, Haotian and Chen, Qianglong and Peng, Weihua and Feng, Xiaocheng and Qin, Bing and Liu, Ting},
title = {A Survey on Hallucination in Large Language Models: Principles, Taxonomy, Challenges, and Open Questions},
year = {2025},
publisher = {Association for Computing Machinery},
volume = {43},
number = {2},
issn = {1046-8188},
url = {https://doi.org/10.1145/3703155},
doi = {10.1145/3703155},
journal = {ACM Transactions on Information Systems},
articleno = {42},
numpages = {55},
}

@article{kalai2025languagemodelshallucinate,
      title={Why Language Models Hallucinate}, 
      author={Adam Tauman Kalai and Ofir Nachum and Santosh S. Vempala and Edwin Zhang},
      year={2025},
      journal={Computing Research Repository},
      volume={arXiv:2509.04664},
      url={https://arxiv.org/abs/2509.04664}, 
}

@article{gao2024retrievalaugmentedgenerationlargelanguage,
      title={Retrieval-Augmented Generation for Large Language Models: A Survey}, 
      author={Yunfan Gao and Yun Xiong and Xinyu Gao and Kangxiang Jia and Jinliu Pan and Yuxi Bi and Yi Dai and Jiawei Sun and Meng Wang and Haofen Wang},
      year={2024},
      url={https://arxiv.org/abs/2312.10997}, 
      volume={arXiv:2312.10997},
      journal={Computing Research Repository},
}

@article{Rackauckas_2024,
   title={Rag-Fusion: A New Take on Retrieval Augmented Generation},
   volume={13},
   ISSN={2319-4111},
   url={http://dx.doi.org/10.5121/ijnlc.2024.13103},
   DOI={10.5121/ijnlc.2024.13103},
   number={1},
   journal={International Journal on Natural Language Computing},
   publisher={Academy and Industry Research Collaboration Center (AIRCC)},
   author={Rackauckas, Zackary},
   year={2024},
   month=feb, pages={37–47} }

@inproceedings{Malinin2021UncertaintyEI,
  title={Uncertainty Estimation in Autoregressive Structured Prediction},
  author={Andrey Malinin and Mark John Francis Gales},
  booktitle={Proceedings of the 2021 International Conference on Learning Representations},
  year={2021},
  url={https://api.semanticscholar.org/CorpusID:231895728}
}

@inproceedings{azaria-mitchell-2023-internal,
    title = {The Internal State of an {LLM} Knows When It`s Lying},
    author = {Azaria, Amos  and
      Mitchell, Tom},
    booktitle = {Proceedings of the 2023 Conference on Empirical Methods in Natural Language Processing},
    year = {2023},
    publisher = {Association for Computational Linguistics},
    url = {https://aclanthology.org/2023.findings-emnlp.68/},
    doi = {10.18653/v1/2023.findings-emnlp.68},
    pages = {967--976},
}

@article{fan2026halluhardhardmultiturnhallucination,
                            title={HalluHard: A Hard Multi-Turn Hallucination Benchmark}, 
                            author={Dongyang Fan and Sebastien Delsad and Nicolas Flammarion and Maksym Andriushchenko},
                            volume={arXiv:2602.01031},
                            journal={Computing Research Repository},
                            year={2026},
                            url={https://arxiv.org/abs/2602.01031}, 
                      }

@article{yang2025qwen3technicalreport,
      title={Qwen3 Technical Report}, 
      author={An Yang and Anfeng Li and Baosong Yang and Beichen Zhang and Binyuan Hui and Bo Zheng and Bowen Yu and Chang Gao and Chengen Huang and Chenxu Lv and Chujie Zheng and Dayiheng Liu and Fan Zhou and Fei Huang and Feng Hu and Hao Ge and Haoran Wei and Huan Lin and Jialong Tang and Jian Yang and Jianhong Tu and Jianwei Zhang and Jianxin Yang and Jiaxi Yang and Jing Zhou and Jingren Zhou and Junyang Lin and Kai Dang and Keqin Bao and Kexin Yang and Le Yu and Lianghao Deng and Mei Li and Mingfeng Xue and Mingze Li and Pei Zhang and Peng Wang and Qin Zhu and Rui Men and Ruize Gao and Shixuan Liu and Shuang Luo and Tianhao Li and Tianyi Tang and Wenbiao Yin and Xingzhang Ren and Xinyu Wang and Xinyu Zhang and Xuancheng Ren and Yang Fan and Yang Su and Yichang Zhang and Yinger Zhang and Yu Wan and Yuqiong Liu and Zekun Wang and Zeyu Cui and Zhenru Zhang and Zhipeng Zhou and Zihan Qiu},
      year={2025},
      journal={Computing Research Repository},
      volume={arXiv:2505.09388},
      url={https://arxiv.org/abs/2505.09388}, 
}

@misc{mistralai2023mistral7binstructv02,
  title = {{Mistral-7B-Instruct-v0.2}},
  author = {{Mistral AI}},
  year = {2023},
  url = {https://huggingface.co/mistralai/Mistral-7B-Instruct-v0.2},
  note = {Hugging Face model card}
}

@article{grattafiori2024llama3herdmodels,
      title={The Llama 3 Herd of Models}, 
      author={Aaron Grattafiori and Abhimanyu Dubey and Abhinav Jauhri and Abhinav Pandey and Abhishek Kadian and Ahmad Al-Dahle and Aiesha Letman and Akhil Mathur and Alan Schelten and Alex Vaughan and Amy Yang and Angela Fan and Anirudh Goyal and Anthony Hartshorn and Aobo Yang and Archi Mitra and Archie Sravankumar and Artem Korenev and Arthur Hinsvark and Arun Rao and Aston Zhang and Aurelien Rodriguez and Austen Gregerson and Ava Spataru and Baptiste Roziere and Bethany Biron and Binh Tang and Bobbie Chern and Charlotte Caucheteux and Chaya Nayak and Chloe Bi and Chris Marra and Chris McConnell and Christian Keller and Christophe Touret and Chunyang Wu and Corinne Wong and Cristian Canton Ferrer and Cyrus Nikolaidis and Damien Allonsius and Daniel Song and Danielle Pintz and Danny Livshits and Danny Wyatt and David Esiobu and Dhruv Choudhary and Dhruv Mahajan and Diego Garcia-Olano and Diego Perino and Dieuwke Hupkes and Egor Lakomkin and Ehab AlBadawy and Elina Lobanova and Emily Dinan and Eric Michael Smith and Filip Radenovic and Francisco Guzmán and Frank Zhang and Gabriel Synnaeve and Gabrielle Lee and Georgia Lewis Anderson and Govind Thattai and Graeme Nail and Gregoire Mialon and Guan Pang and Guillem Cucurell and Hailey Nguyen and Hannah Korevaar and Hu Xu and Hugo Touvron and Iliyan Zarov and Imanol Arrieta Ibarra and Isabel Kloumann and Ishan Misra and Ivan Evtimov and Jack Zhang and Jade Copet and Jaewon Lee and Jan Geffert and Jana Vranes and Jason Park and Jay Mahadeokar and Jeet Shah and Jelmer van der Linde and Jennifer Billock and Jenny Hong and Jenya Lee and Jeremy Fu and Jianfeng Chi and Jianyu Huang and Jiawen Liu and Jie Wang and Jiecao Yu and Joanna Bitton and Joe Spisak and Jongsoo Park and Joseph Rocca and Joshua Johnstun and Joshua Saxe and Junteng Jia and Kalyan Vasuden Alwala and Karthik Prasad and Kartikeya Upasani and Kate Plawiak and Ke Li and Kenneth Heafield and Kevin Stone and Khalid El-Arini and Krithika Iyer and Kshitiz Malik and Kuenley Chiu and Kunal Bhalla and Kushal Lakhotia and Lauren Rantala-Yeary and Laurens van der Maaten and Lawrence Chen and Liang Tan and Liz Jenkins and Louis Martin and Lovish Madaan and Lubo Malo and Lukas Blecher and Lukas Landzaat and Luke de Oliveira and Madeline Muzzi and Mahesh Pasupuleti and Mannat Singh and Manohar Paluri and Marcin Kardas and Maria Tsimpoukelli and Mathew Oldham and Mathieu Rita and Maya Pavlova and Melanie Kambadur and Mike Lewis and Min Si and Mitesh Kumar Singh and Mona Hassan and Naman Goyal and Narjes Torabi and Nikolay Bashlykov and Nikolay Bogoychev and Niladri Chatterji and Ning Zhang and Olivier Duchenne and Onur Çelebi and Patrick Alrassy and Pengchuan Zhang and Pengwei Li and Petar Vasic and Peter Weng and Prajjwal Bhargava and Pratik Dubal and Praveen Krishnan and Punit Singh Koura and Puxin Xu and Qing He and Qingxiao Dong and Ragavan Srinivasan and Raj Ganapathy and Ramon Calderer and Ricardo Silveira Cabral and Robert Stojnic and Roberta Raileanu and Rohan Maheswari and Rohit Girdhar and Rohit Patel and Romain Sauvestre and Ronnie Polidoro and Roshan Sumbaly and Ross Taylor and Ruan Silva and Rui Hou and Rui Wang and Saghar Hosseini and Sahana Chennabasappa and Sanjay Singh and Sean Bell and Seohyun Sonia Kim and Sergey Edunov and Shaoliang Nie and Sharan Narang and Sharath Raparthy and Sheng Shen and Shengye Wan and Shruti Bhosale and Shun Zhang and Simon Vandenhende and Soumya Batra and Spencer Whitman and Sten Sootla and Stephane Collot and Suchin Gururangan and Sydney Borodinsky and Tamar Herman and Tara Fowler and Tarek Sheasha and Thomas Georgiou and Thomas Scialom and Tobias Speckbacher and Todor Mihaylov and Tong Xiao and Ujjwal Karn and Vedanuj Goswami and Vibhor Gupta and Vignesh Ramanathan and Viktor Kerkez and Vincent Gonguet and Virginie Do and Vish Vogeti and Vítor Albiero and Vladan Petrovic and Weiwei Chu and Wenhan Xiong and Wenyin Fu and Whitney Meers and Xavier Martinet and Xiaodong Wang and Xiaofang Wang and Xiaoqing Ellen Tan and Xide Xia and Xinfeng Xie and Xuchao Jia and Xuewei Wang and Yaelle Goldschlag and Yashesh Gaur and Yasmine Babaei and Yi Wen and Yiwen Song and Yuchen Zhang and Yue Li and Yuning Mao and Zacharie Delpierre Coudert and Zheng Yan and Zhengxing Chen and Zoe Papakipos and Aaditya Singh and Aayushi Srivastava and Abha Jain and Adam Kelsey and Adam Shajnfeld and Adithya Gangidi and Adolfo Victoria and Ahuva Goldstand and Ajay Menon and Ajay Sharma and Alex Boesenberg and Alexei Baevski and Allie Feinstein and Amanda Kallet and Amit Sangani and Amos Teo and Anam Yunus and Andrei Lupu and Andres Alvarado and Andrew Caples and Andrew Gu and Andrew Ho and Andrew Poulton and Andrew Ryan and Ankit Ramchandani and Annie Dong and Annie Franco and Anuj Goyal and Aparajita Saraf and Arkabandhu Chowdhury and Ashley Gabriel and Ashwin Bharambe and Assaf Eisenman and Azadeh Yazdan and Beau James and Ben Maurer and Benjamin Leonhardi and Bernie Huang and Beth Loyd and Beto De Paola and Bhargavi Paranjape and Bing Liu and Bo Wu and Boyu Ni and Braden Hancock and Bram Wasti and Brandon Spence and Brani Stojkovic and Brian Gamido and Britt Montalvo and Carl Parker and Carly Burton and Catalina Mejia and Ce Liu and Changhan Wang and Changkyu Kim and Chao Zhou and Chester Hu and Ching-Hsiang Chu and Chris Cai and Chris Tindal and Christoph Feichtenhofer and Cynthia Gao and Damon Civin and Dana Beaty and Daniel Kreymer and Daniel Li and David Adkins and David Xu and Davide Testuggine and Delia David and Devi Parikh and Diana Liskovich and Didem Foss and Dingkang Wang and Duc Le and Dustin Holland and Edward Dowling and Eissa Jamil and Elaine Montgomery and Eleonora Presani and Emily Hahn and Emily Wood and Eric-Tuan Le and Erik Brinkman and Esteban Arcaute and Evan Dunbar and Evan Smothers and Fei Sun and Felix Kreuk and Feng Tian and Filippos Kokkinos and Firat Ozgenel and Francesco Caggioni and Frank Kanayet and Frank Seide and Gabriela Medina Florez and Gabriella Schwarz and Gada Badeer and Georgia Swee and Gil Halpern and Grant Herman and Grigory Sizov and Guangyi and Zhang and Guna Lakshminarayanan and Hakan Inan and Hamid Shojanazeri and Han Zou and Hannah Wang and Hanwen Zha and Haroun Habeeb and Harrison Rudolph and Helen Suk and Henry Aspegren and Hunter Goldman and Hongyuan Zhan and Ibrahim Damlaj and Igor Molybog and Igor Tufanov and Ilias Leontiadis and Irina-Elena Veliche and Itai Gat and Jake Weissman and James Geboski and James Kohli and Janice Lam and Japhet Asher and Jean-Baptiste Gaya and Jeff Marcus and Jeff Tang and Jennifer Chan and Jenny Zhen and Jeremy Reizenstein and Jeremy Teboul and Jessica Zhong and Jian Jin and Jingyi Yang and Joe Cummings and Jon Carvill and Jon Shepard and Jonathan McPhie and Jonathan Torres and Josh Ginsburg and Junjie Wang and Kai Wu and Kam Hou U and Karan Saxena and Kartikay Khandelwal and Katayoun Zand and Kathy Matosich and Kaushik Veeraraghavan and Kelly Michelena and Keqian Li and Kiran Jagadeesh and Kun Huang and Kunal Chawla and Kyle Huang and Lailin Chen and Lakshya Garg and Lavender A and Leandro Silva and Lee Bell and Lei Zhang and Liangpeng Guo and Licheng Yu and Liron Moshkovich and Luca Wehrstedt and Madian Khabsa and Manav Avalani and Manish Bhatt and Martynas Mankus and Matan Hasson and Matthew Lennie and Matthias Reso and Maxim Groshev and Maxim Naumov and Maya Lathi and Meghan Keneally and Miao Liu and Michael L. Seltzer and Michal Valko and Michelle Restrepo and Mihir Patel and Mik Vyatskov and Mikayel Samvelyan and Mike Clark and Mike Macey and Mike Wang and Miquel Jubert Hermoso and Mo Metanat and Mohammad Rastegari and Munish Bansal and Nandhini Santhanam and Natascha Parks and Natasha White and Navyata Bawa and Nayan Singhal and Nick Egebo and Nicolas Usunier and Nikhil Mehta and Nikolay Pavlovich Laptev and Ning Dong and Norman Cheng and Oleg Chernoguz and Olivia Hart and Omkar Salpekar and Ozlem Kalinli and Parkin Kent and Parth Parekh and Paul Saab and Pavan Balaji and Pedro Rittner and Philip Bontrager and Pierre Roux and Piotr Dollar and Polina Zvyagina and Prashant Ratanchandani and Pritish Yuvraj and Qian Liang and Rachad Alao and Rachel Rodriguez and Rafi Ayub and Raghotham Murthy and Raghu Nayani and Rahul Mitra and Rangaprabhu Parthasarathy and Raymond Li and Rebekkah Hogan and Robin Battey and Rocky Wang and Russ Howes and Ruty Rinott and Sachin Mehta and Sachin Siby and Sai Jayesh Bondu and Samyak Datta and Sara Chugh and Sara Hunt and Sargun Dhillon and Sasha Sidorov and Satadru Pan and Saurabh Mahajan and Saurabh Verma and Seiji Yamamoto and Sharadh Ramaswamy and Shaun Lindsay and Shaun Lindsay and Sheng Feng and Shenghao Lin and Shengxin Cindy Zha and Shishir Patil and Shiva Shankar and Shuqiang Zhang and Shuqiang Zhang and Sinong Wang and Sneha Agarwal and Soji Sajuyigbe and Soumith Chintala and Stephanie Max and Stephen Chen and Steve Kehoe and Steve Satterfield and Sudarshan Govindaprasad and Sumit Gupta and Summer Deng and Sungmin Cho and Sunny Virk and Suraj Subramanian and Sy Choudhury and Sydney Goldman and Tal Remez and Tamar Glaser and Tamara Best and Thilo Koehler and Thomas Robinson and Tianhe Li and Tianjun Zhang and Tim Matthews and Timothy Chou and Tzook Shaked and Varun Vontimitta and Victoria Ajayi and Victoria Montanez and Vijai Mohan and Vinay Satish Kumar and Vishal Mangla and Vlad Ionescu and Vlad Poenaru and Vlad Tiberiu Mihailescu and Vladimir Ivanov and Wei Li and Wenchen Wang and Wenwen Jiang and Wes Bouaziz and Will Constable and Xiaocheng Tang and Xiaojian Wu and Xiaolan Wang and Xilun Wu and Xinbo Gao and Yaniv Kleinman and Yanjun Chen and Ye Hu and Ye Jia and Ye Qi and Yenda Li and Yilin Zhang and Ying Zhang and Yossi Adi and Youngjin Nam and Yu and Wang and Yu Zhao and Yuchen Hao and Yundi Qian and Yunlu Li and Yuzi He and Zach Rait and Zachary DeVito and Zef Rosnbrick and Zhaoduo Wen and Zhenyu Yang and Zhiwei Zhao and Zhiyu Ma},
      year={2024},
      journal={Computing Research Repository},
      volume={arXiv:2407.21783},
      url={https://arxiv.org/abs/2407.21783}, 
}

@inproceedings{niu-etal-2024-ragtruth,
    title = {{RAGT}ruth: A Hallucination Corpus for Developing Trustworthy Retrieval-Augmented Language Models},
    author = {Niu, Cheng  and
      Wu, Yuanhao  and
      Zhu, Juno  and
      Xu, Siliang  and
      Shum, KaShun  and
      Zhong, Randy  and
      Song, Juntong  and
      Zhang, Tong},
    booktitle = {Proceedings of the 62nd Annual Meeting of the Association for Computational Linguistics (Volume 1: Long Papers)},
    year = {2024},
    url = "https://aclanthology.org/2024.acl-long.585/",
    publisher = {Association for Computational Linguistics},
    pages = {10862--10878},
}

@article{ridder2024halluragdatasetdetectingcloseddomain,
      title={The HalluRAG Dataset: Detecting Closed-Domain Hallucinations in RAG Applications Using an LLM's Internal States}, 
      journal={Computing Research Repository},
      author={Fabian Ridder and Malte Schilling},
      journal={Computing Research Repository},
      year={2024},
      volume={arXiv:2412.17056},
      url={https://arxiv.org/abs/2412.17056}, 
}

@inproceedings{furumai-etal-2024-zero,
    title = {Zero-shot Persuasive Chatbots with {LLM}-Generated Strategies and Information Retrieval},
    author = {Furumai, Kazuaki  and
      Legaspi, Roberto  and
      Romero, Julio Cesar Vizcarra  and
      Yamazaki, Yudai  and
      Nishimura, Yasutaka  and
      Semnani, Sina  and
      Ikeda, Kazushi  and
      Shi, Weiyan  and
      Lam, Monica},
    booktitle = {Proceedings of the 2024 Conference on Empirical Methods in Natural Language Processing},
    year = {2024},
    publisher = {Association for Computational Linguistics},
    url = {https://aclanthology.org/2024.findings-emnlp.656/},
    doi = {10.18653/v1/2024.findings-emnlp.656},
    pages = {11224--11249},
}

@inproceedings{
    xiong2026toward,
    title={Toward Faithful Retrieval-Augmented Generation with Sparse Autoencoders},
    author={Guangzhi Xiong and Zhenghao He and Bohan Liu and Sanchit Sinha and Aidong Zhang},
    booktitle={The Fourteenth International Conference on Learning Representations},
    year={2026},
    url={https://openreview.net/forum?id=hgBZP67BkP}
}

@article{rabiner1989tutorial,
author = {Rabiner, Lawrence R.},
title = {A tutorial on hidden Markov models and selected applications in speech recognition},
year = {1990},
isbn = {1558601244},
publisher = {Morgan Kaufmann Publishers Inc.},
booktitle = {Readings in Speech Recognition},
pages = {267–296},
numpages = {30}
}

@inproceedings{ganesh-etal-2026-rethinking,
    title = "Rethinking Hallucinations: Correctness, Consistency, and Prompt Multiplicity",
    author = "Ganesh, Prakhar  and
      Shokri, Reza  and
      Farnadi, Golnoosh",
    editor = "Demberg, Vera  and
      Inui, Kentaro  and
      Marquez, Llu{\'i}s",
    booktitle = "Proceedings of the 19th Conference of the {E}uropean Chapter of the {A}ssociation for {C}omputational {L}inguistics (Volume 1: Long Papers)",
    year = "2026",
    address = "Rabat, Morocco",
    publisher = "Association for Computational Linguistics",
    url = "https://aclanthology.org/2026.eacl-long.327/",
    doi = "10.18653/v1/2026.eacl-long.327",
    pages = "6959--6978",
    ISBN = "979-8-89176-380-7"
}

@inproceedings{NEURIPS2020_6b493230_rag,
 author = {Lewis, Patrick and Perez, Ethan and Piktus, Aleksandra and Petroni, Fabio and Karpukhin, Vladimir and Goyal, Naman and K\"{u}ttler, Heinrich and Lewis, Mike and Yih, Wen-tau and Rockt\"{a}schel, Tim and Riedel, Sebastian and Kiela, Douwe},
 booktitle = {Advances in Neural Information Processing Systems},
 editor = {H. Larochelle and M. Ranzato and R. Hadsell and M.F. Balcan and H. Lin},
 pages = {9459--9474},
 publisher = {Curran Associates, Inc.},
 title = {Retrieval-Augmented Generation for Knowledge-Intensive NLP Tasks},
 url = {https://proceedings.neurips.cc/paper_files/paper/2020/file/6b493230205f780e1bc26945df7481e5-Paper.pdf},
 volume = {33},
 year = {2020}
}

@inproceedings{liu-etal-2025-attention,
    title = "Attention-guided Self-reflection for Zero-shot Hallucination Detection in Large Language Models",
    author = "Liu, Qiang  and
      Chen, Xinlong  and
      Ding, Yue  and
      Song, Bowen  and
      Wang, Weiqiang  and
      Wu, Shu  and
      Wang, Liang",
    editor = "Christodoulopoulos, Christos  and
      Chakraborty, Tanmoy  and
      Rose, Carolyn  and
      Peng, Violet",
    booktitle = "Proceedings of the 2025 Conference on Empirical Methods in Natural Language Processing",
    year = "2025",
    publisher = "Association for Computational Linguistics",
    url = "https://aclanthology.org/2025.emnlp-main.1063/",
    doi = "10.18653/v1/2025.emnlp-main.1063",
    pages = "21005--21021",
    ISBN = "979-8-89176-332-6"
}

@inproceedings{bang-etal-2025-hallulens,
    title = "{H}allu{L}ens: {LLM} Hallucination Benchmark",
    author = "Bang, Yejin  and
      Ji, Ziwei  and
      Schelten, Alan  and
      Hartshorn, Anthony  and
      Fowler, Tara  and
      Zhang, Cheng  and
      Cancedda, Nicola  and
      Fung, Pascale",
    editor = "Che, Wanxiang  and
      Nabende, Joyce  and
      Shutova, Ekaterina  and
      Pilehvar, Mohammad Taher",
    booktitle = "Proceedings of the 63rd Annual Meeting of the Association for Computational Linguistics (Volume 1: Long Papers)",
    year = "2025",
    publisher = "Association for Computational Linguistics",
    url = "https://aclanthology.org/2025.acl-long.1176/",
    doi = "10.18653/v1/2025.acl-long.1176",
    pages = "24128--24156",
    ISBN = "979-8-89176-251-0",
}

@misc{openai2024gpt4ocard,
      title={GPT-4o System Card}, 
      author={OpenAI and : and Aaron Hurst and Adam Lerer and Adam P. Goucher and Adam Perelman and Aditya Ramesh and Aidan Clark and AJ Ostrow and Akila Welihinda and Alan Hayes and Alec Radford and Aleksander Mądry and Alex Baker-Whitcomb and Alex Beutel and Alex Borzunov and Alex Carney and Alex Chow and Alex Kirillov and Alex Nichol and Alex Paino and Alex Renzin and Alex Tachard Passos and Alexander Kirillov and Alexi Christakis and Alexis Conneau and Ali Kamali and Allan Jabri and Allison Moyer and Allison Tam and Amadou Crookes and Amin Tootoochian and Amin Tootoonchian and Ananya Kumar and Andrea Vallone and Andrej Karpathy and Andrew Braunstein and Andrew Cann and Andrew Codispoti and Andrew Galu and Andrew Kondrich and Andrew Tulloch and Andrey Mishchenko and Angela Baek and Angela Jiang and Antoine Pelisse and Antonia Woodford and Anuj Gosalia and Arka Dhar and Ashley Pantuliano and Avi Nayak and Avital Oliver and Barret Zoph and Behrooz Ghorbani and Ben Leimberger and Ben Rossen and Ben Sokolowsky and Ben Wang and Benjamin Zweig and Beth Hoover and Blake Samic and Bob McGrew and Bobby Spero and Bogo Giertler and Bowen Cheng and Brad Lightcap and Brandon Walkin and Brendan Quinn and Brian Guarraci and Brian Hsu and Bright Kellogg and Brydon Eastman and Camillo Lugaresi and Carroll Wainwright and Cary Bassin and Cary Hudson and Casey Chu and Chad Nelson and Chak Li and Chan Jun Shern and Channing Conger and Charlotte Barette and Chelsea Voss and Chen Ding and Cheng Lu and Chong Zhang and Chris Beaumont and Chris Hallacy and Chris Koch and Christian Gibson and Christina Kim and Christine Choi and Christine McLeavey and Christopher Hesse and Claudia Fischer and Clemens Winter and Coley Czarnecki and Colin Jarvis and Colin Wei and Constantin Koumouzelis and Dane Sherburn and Daniel Kappler and Daniel Levin and Daniel Levy and David Carr and David Farhi and David Mely and David Robinson and David Sasaki and Denny Jin and Dev Valladares and Dimitris Tsipras and Doug Li and Duc Phong Nguyen and Duncan Findlay and Edede Oiwoh and Edmund Wong and Ehsan Asdar and Elizabeth Proehl and Elizabeth Yang and Eric Antonow and Eric Kramer and Eric Peterson and Eric Sigler and Eric Wallace and Eugene Brevdo and Evan Mays and Farzad Khorasani and Felipe Petroski Such and Filippo Raso and Francis Zhang and Fred von Lohmann and Freddie Sulit and Gabriel Goh and Gene Oden and Geoff Salmon and Giulio Starace and Greg Brockman and Hadi Salman and Haiming Bao and Haitang Hu and Hannah Wong and Haoyu Wang and Heather Schmidt and Heather Whitney and Heewoo Jun and Hendrik Kirchner and Henrique Ponde de Oliveira Pinto and Hongyu Ren and Huiwen Chang and Hyung Won Chung and Ian Kivlichan and Ian O'Connell and Ian O'Connell and Ian Osband and Ian Silber and Ian Sohl and Ibrahim Okuyucu and Ikai Lan and Ilya Kostrikov and Ilya Sutskever and Ingmar Kanitscheider and Ishaan Gulrajani and Jacob Coxon and Jacob Menick and Jakub Pachocki and James Aung and James Betker and James Crooks and James Lennon and Jamie Kiros and Jan Leike and Jane Park and Jason Kwon and Jason Phang and Jason Teplitz and Jason Wei and Jason Wolfe and Jay Chen and Jeff Harris and Jenia Varavva and Jessica Gan Lee and Jessica Shieh and Ji Lin and Jiahui Yu and Jiayi Weng and Jie Tang and Jieqi Yu and Joanne Jang and Joaquin Quinonero Candela and Joe Beutler and Joe Landers and Joel Parish and Johannes Heidecke and John Schulman and Jonathan Lachman and Jonathan McKay and Jonathan Uesato and Jonathan Ward and Jong Wook Kim and Joost Huizinga and Jordan Sitkin and Jos Kraaijeveld and Josh Gross and Josh Kaplan and Josh Snyder and Joshua Achiam and Joy Jiao and Joyce Lee and Juntang Zhuang and Justyn Harriman and Kai Fricke and Kai Hayashi and Karan Singhal and Katy Shi and Kavin Karthik and Kayla Wood and Kendra Rimbach and Kenny Hsu and Kenny Nguyen and Keren Gu-Lemberg and Kevin Button and Kevin Liu and Kiel Howe and Krithika Muthukumar and Kyle Luther and Lama Ahmad and Larry Kai and Lauren Itow and Lauren Workman and Leher Pathak and Leo Chen and Li Jing and Lia Guy and Liam Fedus and Liang Zhou and Lien Mamitsuka and Lilian Weng and Lindsay McCallum and Lindsey Held and Long Ouyang and Louis Feuvrier and Lu Zhang and Lukas Kondraciuk and Lukasz Kaiser and Luke Hewitt and Luke Metz and Lyric Doshi and Mada Aflak and Maddie Simens and Madelaine Boyd and Madeleine Thompson and Marat Dukhan and Mark Chen and Mark Gray and Mark Hudnall and Marvin Zhang and Marwan Aljubeh and Mateusz Litwin and Matthew Zeng and Max Johnson and Maya Shetty and Mayank Gupta and Meghan Shah and Mehmet Yatbaz and Meng Jia Yang and Mengchao Zhong and Mia Glaese and Mianna Chen and Michael Janner and Michael Lampe and Michael Petrov and Michael Wu and Michele Wang and Michelle Fradin and Michelle Pokrass and Miguel Castro and Miguel Oom Temudo de Castro and Mikhail Pavlov and Miles Brundage and Miles Wang and Minal Khan and Mira Murati and Mo Bavarian and Molly Lin and Murat Yesildal and Nacho Soto and Natalia Gimelshein and Natalie Cone and Natalie Staudacher and Natalie Summers and Natan LaFontaine and Neil Chowdhury and Nick Ryder and Nick Stathas and Nick Turley and Nik Tezak and Niko Felix and Nithanth Kudige and Nitish Keskar and Noah Deutsch and Noel Bundick and Nora Puckett and Ofir Nachum and Ola Okelola and Oleg Boiko and Oleg Murk and Oliver Jaffe and Olivia Watkins and Olivier Godement and Owen Campbell-Moore and Patrick Chao and Paul McMillan and Pavel Belov and Peng Su and Peter Bak and Peter Bakkum and Peter Deng and Peter Dolan and Peter Hoeschele and Peter Welinder and Phil Tillet and Philip Pronin and Philippe Tillet and Prafulla Dhariwal and Qiming Yuan and Rachel Dias and Rachel Lim and Rahul Arora and Rajan Troll and Randall Lin and Rapha Gontijo Lopes and Raul Puri and Reah Miyara and Reimar Leike and Renaud Gaubert and Reza Zamani and Ricky Wang and Rob Donnelly and Rob Honsby and Rocky Smith and Rohan Sahai and Rohit Ramchandani and Romain Huet and Rory Carmichael and Rowan Zellers and Roy Chen and Ruby Chen and Ruslan Nigmatullin and Ryan Cheu and Saachi Jain and Sam Altman and Sam Schoenholz and Sam Toizer and Samuel Miserendino and Sandhini Agarwal and Sara Culver and Scott Ethersmith and Scott Gray and Sean Grove and Sean Metzger and Shamez Hermani and Shantanu Jain and Shengjia Zhao and Sherwin Wu and Shino Jomoto and Shirong Wu and Shuaiqi and Xia and Sonia Phene and Spencer Papay and Srinivas Narayanan and Steve Coffey and Steve Lee and Stewart Hall and Suchir Balaji and Tal Broda and Tal Stramer and Tao Xu and Tarun Gogineni and Taya Christianson and Ted Sanders and Tejal Patwardhan and Thomas Cunninghman and Thomas Degry and Thomas Dimson and Thomas Raoux and Thomas Shadwell and Tianhao Zheng and Todd Underwood and Todor Markov and Toki Sherbakov and Tom Rubin and Tom Stasi and Tomer Kaftan and Tristan Heywood and Troy Peterson and Tyce Walters and Tyna Eloundou and Valerie Qi and Veit Moeller and Vinnie Monaco and Vishal Kuo and Vlad Fomenko and Wayne Chang and Weiyi Zheng and Wenda Zhou and Wesam Manassra and Will Sheu and Wojciech Zaremba and Yash Patil and Yilei Qian and Yongjik Kim and Youlong Cheng and Yu Zhang and Yuchen He and Yuchen Zhang and Yujia Jin and Yunxing Dai and Yury Malkov},
      journal={Computing Research Repository},
      year={2024},
      volume={arXiv:2410.21276},
      url={https://arxiv.org/abs/2410.21276}, 
}
\appendix

\section{Details of Label-Persistence Smoothing}
\label{app:label_persistence_smoothing}

This appendix gives the forward--backward recursions and posterior marginal derivation for label-persistence smoothing. 
The method defines non-negative weights over smoothing-label sequences by combining token-level confidence with label persistence between neighboring labels, so that posterior marginals can be computed efficiently without enumerating all $2^T$ possible label sequences.

\subsection{Forward Recursion}

The forward message \(F_i(\ell)\) is the unnormalized total weight of all partial smoothing-label sequences ending with \(z_i=\ell\):

\begin{equation}
\begin{aligned}
F_i(\ell)
&=
\sum_{z_{1:i}:z_i=\ell}
\pi(z_1)
\prod_{j=1}^{i}
\phi_j(z_j)
\\
&\quad\times
\prod_{j=2}^{i}
\rho(z_{j-1},z_j).
\end{aligned}
\end{equation}

The initialization is
\begin{equation}
F_1(\ell)=\pi(\ell)\phi_1(\ell).
\end{equation}

For \(i=2,\ldots,T\), the recursion is obtained by grouping the partial sequences according to the previous label \(\ell'=z_{i-1}\). Starting from the definition,
\begin{equation}
\begin{aligned}
F_i(\ell)
&=
\sum_{z_{1:i}:z_i=\ell}
\pi(z_1)
\prod_{j=1}^{i}
\phi_j(z_j)
\prod_{j=2}^{i}
\rho(z_{j-1},z_j)
\\
&=
\sum_{\ell'\in\{0,1\}}
\sum_{z_{1:i-1}:z_{i-1}=\ell'}
\pi(z_1)
\prod_{j=1}^{i-1}
\phi_j(z_j)
\\
&\quad\times
\prod_{j=2}^{i-1}
\rho(z_{j-1},z_j)
\rho(\ell',\ell)\phi_i(\ell).
\end{aligned}
\end{equation}
The inner sum is exactly \(F_{i-1}(\ell')\). Therefore,
\begin{equation}
F_i(\ell)
=
\phi_i(\ell)
\sum_{\ell'\in\{0,1\}}
F_{i-1}(\ell')\rho(\ell',\ell).
\end{equation}

Since the smoothing label is binary, the recursion can be written
explicitly. Let \(p=p_{\mathrm{stay}}\). Using
\(\phi_i(1)=s_i\) and \(\phi_i(0)=1-s_i\), we obtain
\begin{equation}
F_i(1)
=
s_i
\left[
pF_{i-1}(1)
+
(1-p)F_{i-1}(0)
\right],
\end{equation}
and
\begin{equation}
F_i(0)
=
(1-s_i)
\left[
pF_{i-1}(0)
+
(1-p)F_{i-1}(1)
\right].
\end{equation}
The corresponding initial values are
\begin{equation}
F_1(1)=\frac{1}{2}s_1,
\qquad
F_1(0)=\frac{1}{2}(1-s_1).
\end{equation}

\subsection{Backward Recursion}

The backward message \(B_i(\ell)\) is the unnormalized total weight of all
suffix smoothing-label sequences from positions \(i+1\) to \(T\),
conditioned on \(z_i=\ell\):
\begin{equation}
\begin{aligned}
B_i(\ell)
&=
\sum_{z_{i+1:T}}
\prod_{j=i+1}^{T}
\phi_j(z_j)
\\
&\quad\times
\prod_{j=i+1}^{T}
\rho(z_{j-1},z_j),
\qquad z_i=\ell .
\end{aligned}
\end{equation}
The initialization is
\begin{equation}
B_T(\ell)=1,
\end{equation}
which corresponds to the empty product beyond the final token.

For \(i=T-1,\ldots,1\), the recursion is obtained by grouping the suffix
sequences according to the next label \(\ell'=z_{i+1}\). Starting from
the definition,
\begin{equation}
\begin{aligned}
B_i(\ell)
&=
\sum_{z_{i+1:T}}
\prod_{j=i+1}^{T}
\phi_j(z_j)
\prod_{j=i+1}^{T}
\rho(z_{j-1},z_j)
\\
&=
\sum_{\ell'\in\{0,1\}}
\sum_{z_{i+2:T}}
\phi_{i+1}(\ell')
\rho(\ell,\ell')
\\
&\quad\times
\prod_{j=i+2}^{T}
\phi_j(z_j)
\prod_{j=i+2}^{T}
\rho(z_{j-1},z_j).
\end{aligned}
\end{equation}
The inner sum is exactly \(B_{i+1}(\ell')\). Therefore,
\begin{equation}
B_i(\ell)
=
\sum_{\ell'\in\{0,1\}}
\rho(\ell,\ell')
\phi_{i+1}(\ell')
B_{i+1}(\ell').
\end{equation}

For the binary case, this recursion is equivalently
\begin{equation}
B_i(1)
=
ps_{i+1}B_{i+1}(1)
+
(1-p)(1-s_{i+1})B_{i+1}(0),
\end{equation}
and
\begin{equation}
B_i(0)
=
p(1-s_{i+1})B_{i+1}(0)
+
(1-p)s_{i+1}B_{i+1}(1).
\end{equation}
The terminal values are
\begin{equation}
B_T(1)=1,
\qquad
B_T(0)=1.
\end{equation}

\subsection{Posterior Marginal}

The posterior marginal for smoothing label \(\ell\) at position \(i\) is
\begin{equation}
q_i(\ell)
=
P(z_i=\ell\mid s_{1:T})
=
\sum_{z_{1:T}:z_i=\ell}
P(z_{1:T}\mid s_{1:T}).
\end{equation}
Substituting the normalized sequence weight gives
\begin{equation}
\begin{aligned}
q_i(\ell)
&=
\frac{1}{Z(s_{1:T})}
\sum_{z_{1:T}:z_i=\ell}
\pi(z_1)
\prod_{j=1}^{T}
\phi_j(z_j)
\\
&\quad\times
\prod_{j=2}^{T}
\rho(z_{j-1},z_j).
\end{aligned}
\end{equation}

Because the sequence weight has a first-order linear-chain factorization, fixing \(z_i=\ell\) separates the unnormalized weight into prefix and suffix terms:
\begin{equation}
\begin{aligned}
&\sum_{z_{1:T}:z_i=\ell}
\pi(z_1)
\prod_{j=1}^{T}
\phi_j(z_j)
\prod_{j=2}^{T}
\rho(z_{j-1},z_j)
\\
&\qquad
=
F_i(\ell)B_i(\ell).
\end{aligned}
\end{equation}
Thus,
\begin{equation}
q_i(\ell)
=
\frac{F_i(\ell)B_i(\ell)}{Z(s_{1:T})}.
\end{equation}

The normalizing constant can be decomposed at position \(i\):
\begin{equation}
Z(s_{1:T})
=
\sum_{\ell'\in\{0,1\}}
F_i(\ell')B_i(\ell').
\end{equation}
Therefore,
\begin{equation}
q_i(\ell)
=
\frac{
F_i(\ell)B_i(\ell)
}{
\sum_{\ell'\in\{0,1\}}
F_i(\ell')B_i(\ell')
}.
\end{equation}
The final smoothed hallucination score is
\begin{equation}
\tilde{s}_i=q_i(1)
=
\frac{
F_i(1)B_i(1)
}{
F_i(0)B_i(0)+F_i(1)B_i(1)
}.
\end{equation}

\section{Sensitivity Analysis of \(p_{\mathrm{stay}}\)}
\label{sec:pstay_sensitivity}
\begin{figure*}[t]
  \centering
  \includegraphics[width=\textwidth]{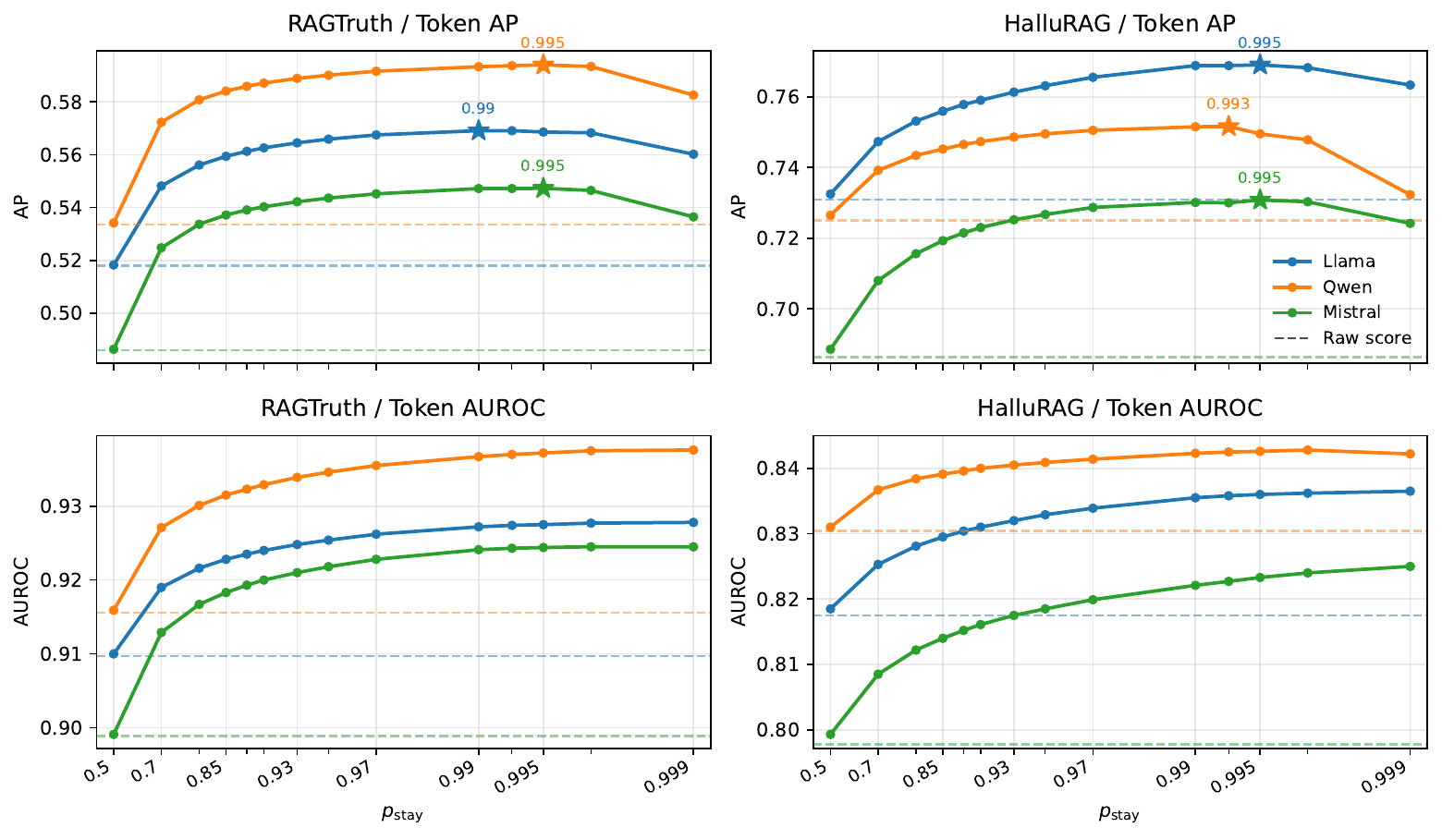}
  \caption{
Sensitivity of token-level hallucination detection performance to \(p_{\mathrm{stay}}\). Dashed horizontal lines indicate raw scores before label-persistence smoothing.
  }
  \label{fig:token_pstay_sensitivity}
\end{figure*}

\begin{figure*}[th]
  \centering
  \includegraphics[width=\textwidth]{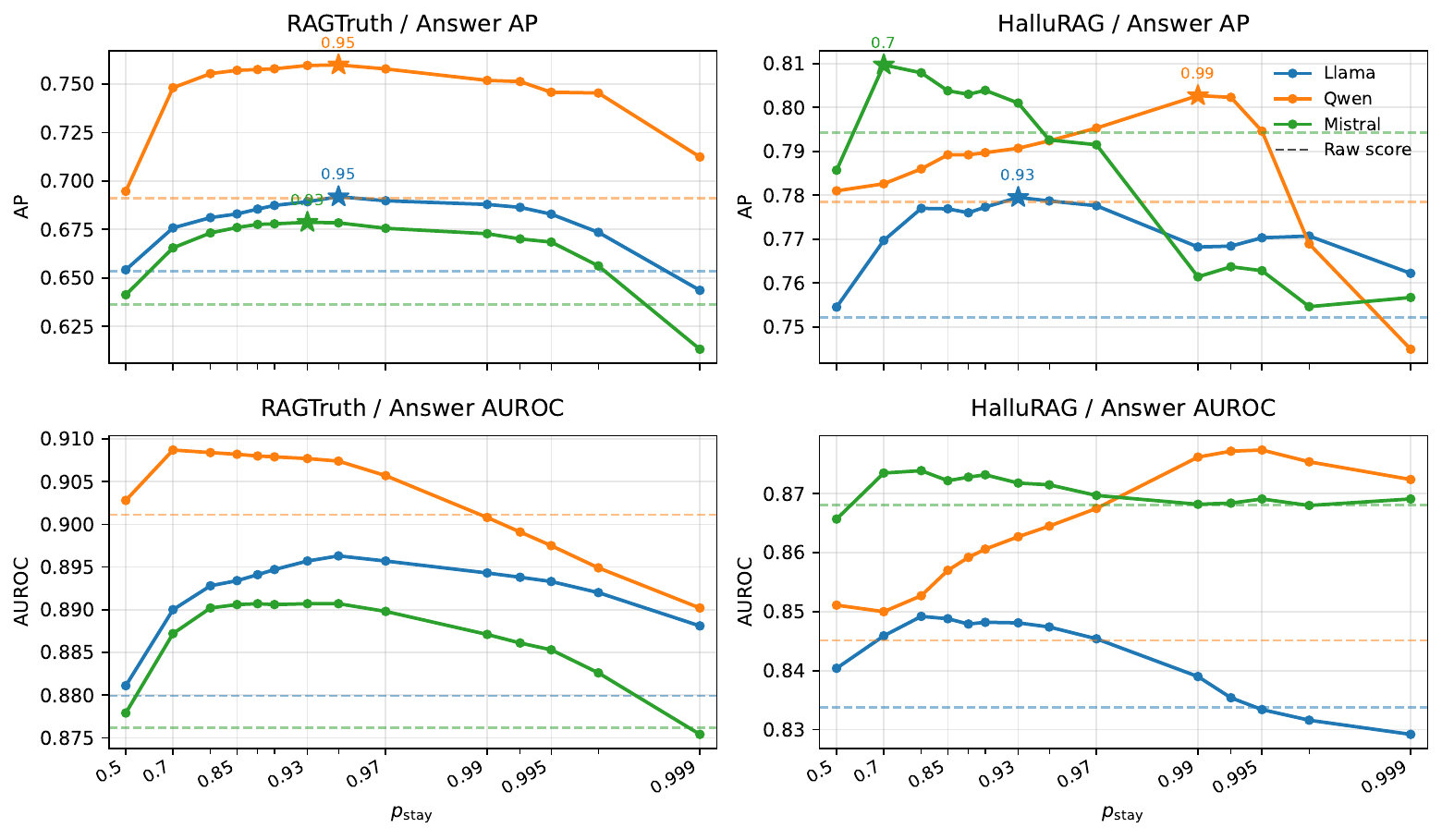}
  \caption{
Sensitivity of answer-level hallucination detection performance to \(p_{\mathrm{stay}}\). Answer-level scores are obtained by taking the maximum over token-level hallucination scores. Dashed horizontal lines indicate raw scores before label-persistence smoothing.
  }
  \label{fig:answer_pstay_sensitivity}
\end{figure*}

We analyze the sensitivity of CORTEX to the self-loop probability
\(p_{\mathrm{stay}}\) used in label-persistence smoothing. 
This parameter controls the strength of span-level smoothing: larger values encourage neighboring tokens to remain in the same smoothing-label variable, whereas smaller values keep the smoothed scores closer to the raw token-level classifier outputs.

Figures~\ref{fig:token_pstay_sensitivity} and~\ref{fig:answer_pstay_sensitivity} show the results for token-level and answer-level evaluation, respectively. 
The dashed horizontal lines indicate the raw scores before label-persistence smoothing. 
Across both datasets and all open-weight LLMs, label-persistence smoothing improves over the raw scores for a wide range of \(p_{\mathrm{stay}}\), indicating that the gains of CORTEX are not tied to a single finely tuned value.

For token-level detection, performance is generally stable when \(p_{\mathrm{stay}}\) is large. 
On both RAGTruth and HalluRAG, AP increases as \(p_{\mathrm{stay}}\) becomes larger and then forms a broad plateau around high values such as \(0.99\), \(0.993\), and \(0.995\). 
AUROC shows a similar trend, with improvements over the raw scores maintained across a wide range of high \(p_{\mathrm{stay}}\) values. 
This behavior is consistent with the role of label-persistence smoothing in token-level hallucination detection: hallucinated content typically appears as contiguous spans, and therefore stronger self-transition probabilities help suppress isolated noisy predictions while preserving coherent hallucinated regions.

For answer-level detection, the optimal \(p_{\mathrm{stay}}\) tends to be smaller than in token-level evaluation. 
This is because answer-level scores are obtained by aggregating token-level scores with the maximum operator. 
If \(p_{\mathrm{stay}}\) is too large, smoothing may overly spread high scores across neighboring tokens or make span boundaries too persistent, which can affect the maximum score used for answer-level classification. 
Nevertheless, label-persistence smoothing remains beneficial across a broad range of values, and the selected setting \(p_{\mathrm{stay}}=0.93\) provides strong and stable answer-level performance.

Based on these results, we use \(p_{\mathrm{stay}}=0.993\) for token-level evaluation and \(p_{\mathrm{stay}}=0.93\) for answer-level evaluation in the main experiments. 
We use a single value for each evaluation granularity rather than tuning \(p_{\mathrm{stay}}\) separately for each dataset and open-weight LLM. 
This setting keeps the evaluation protocol simple while still capturing the main benefit of label-persistence smoothing: converting locally noisy token-level scores into more span-consistent hallucination estimates.

\section{Experimental Details}
\label{sec:experimental_details}

\paragraph{MLP classifier.}
For all methods that require a supervised classifier, including CORTEX and MLP-based baselines, we use the same three-layer MLP architecture to isolate the effect of the input features.
The classifier consists of two hidden linear layers with 256 and 128 hidden units, respectively, followed by a final linear layer that produces a scalar output logit.
Each hidden layer is followed by a ReLU activation and dropout with a rate of 0.1.
We train the classifier using AdamW with a learning rate of $1\times 10^{-3}$ and a weight decay of $1\times 10^{-4}$.
The batch size is set to 4096, and the model is trained for 10 epochs.
The same hyperparameters are used across all datasets, models, and methods unless otherwise stated.

\paragraph{Representation extraction.}
To obtain internal representations from the open-weight analysis model $M_{\mathrm{open}}$, we do not perform autoregressive generation.
Instead, we feed the constructed input sequence to $M_{\mathrm{open}}$ and extract the internal representations computed in a single forward pass.
This implementation matches the post-hoc setting considered in this work: the answer text has already been generated by the closed-weight model, and $M_{\mathrm{open}}$ is used only to analyze the given answer.

\paragraph{Computational cost.}
All experiments are conducted on a single NVIDIA A100 GPU.
CORTEX is lightweight in practice: on RAGTruth, feature extraction, classifier training, and evaluation take approximately 20 minutes in total, while on HalluRAG they take approximately 5 minutes in total.
This indicates that CORTEX incurs only modest computational overhead while providing token-level hallucination scores.

\paragraph{Artifact Licenses and Use.}
We use RAGTruth and HalluRAG as existing public evaluation benchmarks and do not redistribute the datasets.
RAGTruth is released under the MIT License.
HalluRAG is made publicly available by its authors through their repository and dataset DOI, but we did not find an explicit dataset license in the available documentation.
All datasets are used solely for research evaluation, and we cite their original sources.

\paragraph{Dataset Documentation.}
We use only publicly available English-language RAG hallucination benchmarks and do not collect any new data.
Our experiments focus on hallucination detection in generated RAG outputs; we do not use, infer, or analyze demographic attributes.

\section{Effect of Label-persistence Smoothing on Token-Level Baselines}
\label{sec:appendix_smoothing_baselines}

\begin{table*}[t]
\centering
\small
\setlength{\tabcolsep}{3pt}
\renewcommand{\arraystretch}{1.1}

\begin{tabular}{l|cc|cc|cc}
\toprule
\multirow{3}{*}{\textbf{Method}} 
& \multicolumn{6}{c}{\textbf{RAGTruth}} \\
\cmidrule(lr){2-7}

& \multicolumn{2}{c|}{\textbf{Llama}} 
& \multicolumn{2}{c|}{\textbf{Qwen}} 
& \multicolumn{2}{c}{\textbf{Mistral}}  \\
\cmidrule(lr){2-3} \cmidrule(lr){4-5} \cmidrule(lr){6-7}

& AP & AUROC 
& AP & AUROC 
& AP & AUROC \\
\midrule

SAPLMA
& 0.3894 & 0.8879  & 0.4182 & 0.8820 & 0.4027 & 0.8822\\

 {\scriptsize\hspace{2mm}+ Smoothing ($P_{\text{stay}}=0.993$)}
& 0.4910 & 0.9225 & 0.5078 & 0.9163 & 0.4965 & 0.9205  \\

 {\scriptsize\hspace{2mm}+ Smoothing ($P_{\text{stay}}=0.995$)}
& 0.4925 & 0.9228 & 0.5088 & 0.9168 & 0.4972 & 0.9210  \\

 {\scriptsize\hspace{2mm}+ Smoothing ($P_{\text{stay}}=0.997$)}
& 0.4938 & 0.9232 & \underline{0.5089} & 0.9174 & 0.4982 & 0.9216  \\

RAGLens
& 0.4271 & 0.8953 & 0.4575 & 0.9075 & 0.4244 & 0.8884 \\

 {\scriptsize\hspace{2mm}+ Smoothing ($P_{\text{stay}}=0.993$)}
&\underline{0.5191} & 0.9300 & 0.5068 & \underline{0.9286} & 0.5264 & 0.9268 \\

 {\scriptsize\hspace{2mm}+ Smoothing ($P_{\text{stay}}=0.995$)}
& \underline{0.5191} & 0.9302 & 0.5057 & \underline{0.9286} & 0.5276 & 0.9274 \\

 {\scriptsize\hspace{2mm}+ Smoothing ($P_{\text{stay}}=0.997$)}
& \underline{0.5191} & \textbf{0.9305} & 0.5026 & \underline{0.9286} & \underline{0.5290} & \textbf{0.9282} \\

\textbf{CORTEX} 
& 0.5181 & 0.9097 & 0.5335 & 0.9156  & 0.4860 & 0.8989\\

 {\scriptsize\hspace{2mm}+ Smoothing ($P_{\text{stay}}=0.993$)}
& \textbf{0.5691} & 0.9274 & 0.5937 & 0.9370 & 0.5472 & 0.9243 \\

 {\scriptsize\hspace{2mm}+ Smoothing ($P_{\text{stay}}=0.995$)}
& 0.5686 & 0.9275 & \textbf{0.5940} & 0.9372 & \textbf{0.5473} & 0.9244  \\

 {\scriptsize\hspace{2mm}+ Smoothing ($P_{\text{stay}}=0.997$)}
& 0.5683 & \underline{0.9277} & 0.5934 & \textbf{0.9375} & 0.5465 & \underline{0.9245} \\

\bottomrule
\end{tabular}
\caption{Effect of label-persistence smoothing on token-level baselines on RAGTruth. Improvements for both SAPLMA and RAGLens indicate that span-level post-processing is broadly useful.}
\label{tab:appendix_smoothing_baseline_results_ragtruth}
\end{table*}

\begin{table*}[t]
\centering
\small
\setlength{\tabcolsep}{3pt}
\renewcommand{\arraystretch}{1.1}

\begin{tabular}{l|cc|cc|cc}
\toprule
\multirow{3}{*}{\textbf{Method}}  
& \multicolumn{6}{c}{\textbf{HalluRAG}} \\
\cmidrule(lr){2-7}

& \multicolumn{2}{c|}{\textbf{Llama}} 
& \multicolumn{2}{c|}{\textbf{Qwen}} 
& \multicolumn{2}{c}{\textbf{Mistral}} \\
\cmidrule(lr){2-3} \cmidrule(lr){4-5} \cmidrule(lr){6-7}

& AP & AUROC 
& AP & AUROC 
& AP & AUROC \\
\midrule

SAPLMA
& 0.6875 & 0.8024 & 0.6584 & 0.8129 & 0.6193 & 0.7578 \\

 {\scriptsize\hspace{2mm}+ Smoothing ($P_{\text{stay}}=0.993$)}
& 0.7527 & 0.8291 & 0.7125 & 0.8378 & 0.7132 & 0.8108 \\

 {\scriptsize\hspace{2mm}+ Smoothing ($P_{\text{stay}}=0.995$)} 
& 0.7539 & 0.8295 & 0.7119 & 0.8383 & 0.7160 & 0.8122 \\

 {\scriptsize\hspace{2mm}+ Smoothing ($P_{\text{stay}}=0.997$)} 
& \underline{0.7546} & \underline{0.8301} & 0.7128 & \underline{0.8391} & 0.7173 & 0.8140 \\

RAGLens
& 0.5725 & 0.7425 & 0.6380 & 0.7639 & 0.5596 & 0.7393 \\

 {\scriptsize\hspace{2mm}+ Smoothing ($P_{\text{stay}}=0.993$)}
& 0.7146 & 0.8007 & 0.7668 & 0.8138 & 0.7420 & 0.8214 \\

 {\scriptsize\hspace{2mm}+ Smoothing ($P_{\text{stay}}=0.995$)} 
& 0.7169 & 0.8019 & 0.7695 & 0.8148 & 0.7459 & 0.8232 \\

 {\scriptsize\hspace{2mm}+ Smoothing ($P_{\text{stay}}=0.997$)}
& 0.7202 & 0.8035 & \textbf{0.7732} & 0.8163 & \textbf{0.7513} & \textbf{0.8256}  \\

\textbf{CORTEX} 
& 0.7309 & 0.8175 & 0.7250 & 0.8304 & 0.6864 & 0.7978 \\

 {\scriptsize\hspace{2mm}+ Smoothing ($P_{\text{stay}}=0.993$)}
& 0.7688 & 0.8358 & \underline{0.7516} & 0.8425 & 0.7300 & 0.8227 \\

 {\scriptsize\hspace{2mm}+ Smoothing ($P_{\text{stay}}=0.995$)}
& \textbf{0.7690} & 0.8360 & 0.7495 & 0.8426 & \underline{0.7308} & 0.8233 \\

 {\scriptsize\hspace{2mm}+ Smoothing ($P_{\text{stay}}=0.997$)}
& 0.7682 & \textbf{0.8362} & 0.7478 & \textbf{0.8428} & 0.7303 & \underline{0.8240} \\

\bottomrule
\end{tabular}
\caption{Effect of applying smoothing to token-level baselines on HalluRAG. 
Label-persistence smoothing improves all methods substantially, including SAPLMA and RAGLens. }
\label{tab:appendix_smoothing_baseline_results_hallurag}
\end{table*}

Label-persistence smoothing is a post-processing module that can be applied to any method that produces token-level hallucination scores. 
To examine whether the gains of CORTEX are due only to label-persistence smoothing procedure to two strong token-level baselines, SAPLMA and RAGLens, and compare them with CORTEX under the same settings. 
Tables~\ref{tab:appendix_smoothing_baseline_results_ragtruth} and~\ref{tab:appendix_smoothing_baseline_results_hallurag} report the results on RAGTruth and HalluRAG, respectively.

Before applying label-persistence smoothing, CORTEX achieves the best performance in all settings across both datasets and all open-weight LLMs. 
This result indicates that the comparative representation features of CORTEX already provide a stronger token-level signal than the single-view representation features used by the baselines. 
After label-persistence smoothing is applied, the performance of SAPLMA and RAGLens improves substantially, confirming that span-level post-processing is broadly useful for reducing local prediction noise. 
Nevertheless, CORTEX remains strongest in AP across all RAGTruth settings and remains competitive or superior in most HalluRAG settings. 
These results suggest that label-persistence smoothing is beneficial as a generic post-processing step.

In particular, the comparison before smoothing isolates the effect of the underlying token-level scoring function: CORTEX outperforms the baselines without relying on span-level post-processing. 
The comparison after smoothing further shows that CORTEX can benefit from the same generic smoothing procedure while retaining the advantage of its paired reference-conditioned and no-reference representation comparison. 
Thus, label-persistence smoothing and the comparative representation features play complementary roles: the former improves span consistency, whereas the latter provides a stronger reference-grounded hallucination signal.

\section{Additional Heatmap-Based Ablation Case Studies}
\label{sec:appendix_ablation_case_studies}
\begin{figure*}[t]
  \centering
  \includegraphics[width=\textwidth]{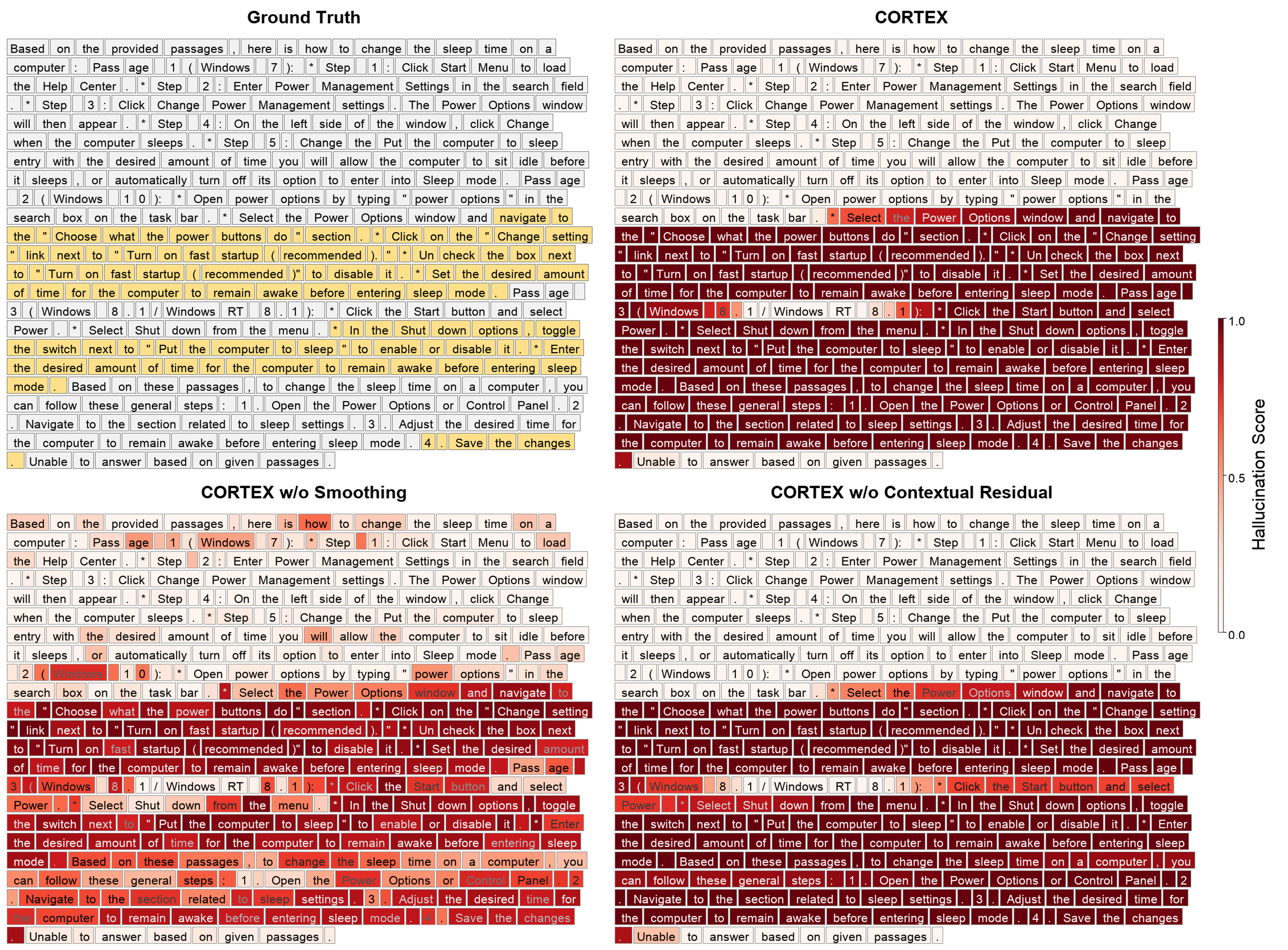}
  \caption{
  Ablation case study with hallucination spans of different granularities. 
  Without label-persistence smoothing, hallucination scores are more fragmented and localized. 
  With label-persistence smoothing, nearby high-score regions are connected into broader span-consistent predictions, although this can also merge distinct hallucinated fragments into a wider continuous region.
  }
  \label{fig:appendix_ablation_various_spans}
\end{figure*}

\begin{figure*}[t]
  \centering
  \includegraphics[width=\textwidth]{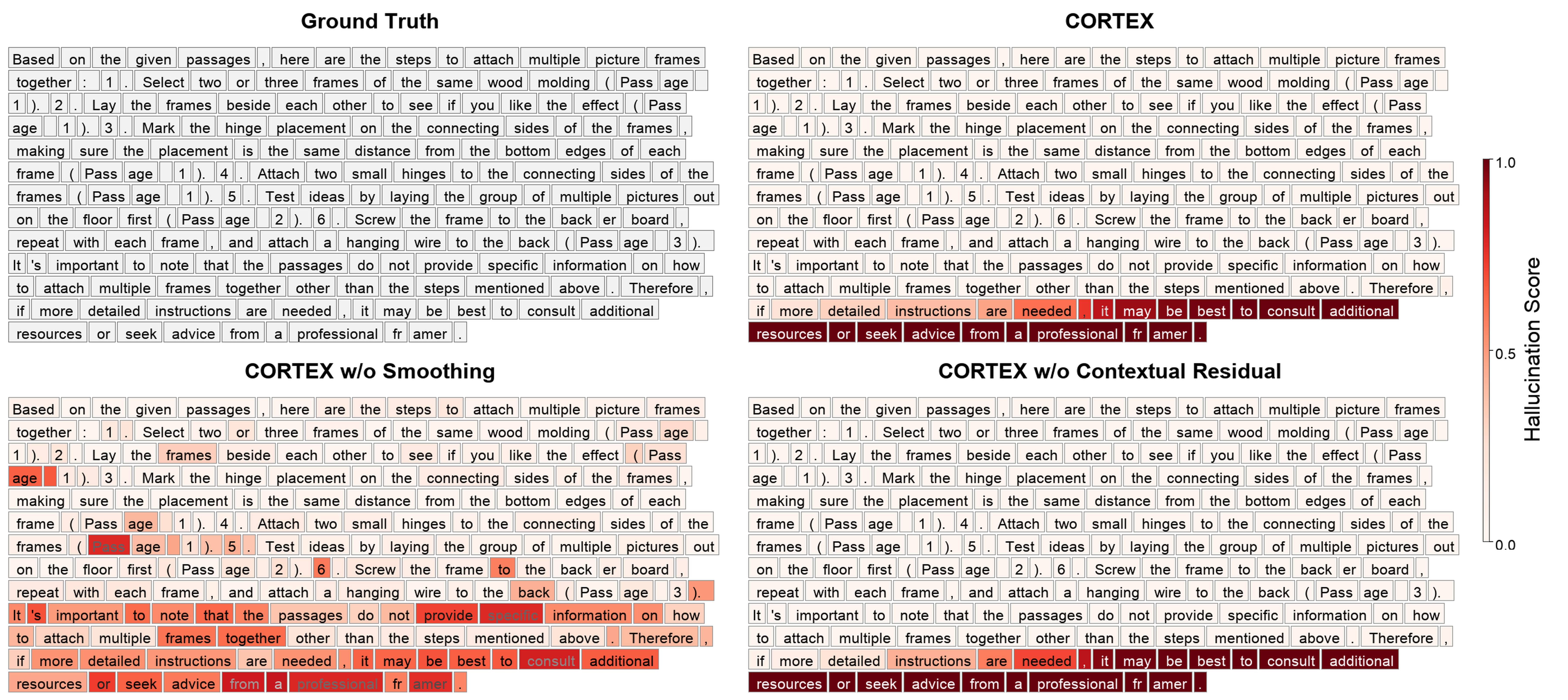}
  \caption{
  Ablation case study involving groundless advice. 
  The answer includes additional advice that may be informative but is not grounded in the provided references. 
  CORTEX assigns high hallucination scores to this region, reflecting its role as a detector of reference support rather than a general factuality or usefulness judge.
  }
  \label{fig:appendix_ablation_advice}
\end{figure*}

We present additional heatmap-based case studies to qualitatively analyze the behavior of each component in CORTEX. 
The heatmaps compare the ground-truth hallucination spans with the predictions of the full CORTEX model, CORTEX without label-persistence smoothing, and CORTEX without the contextual residual \(c\). 
Darker colors indicate higher hallucination scores.

Figure~\ref{fig:appendix_ablation_various_spans} shows an example in which the annotated hallucination spans vary in granularity. 
The ground-truth annotation contains both a broad hallucinated region spanning multiple sentences and more compact hallucinated fragments. 
Without label-persistence smoothing, CORTEX assigns high scores at relatively fine-grained units. 
This behavior suggests that the raw token-level classifier can capture localized hallucination signals, but its predictions are fragmented and locally unstable. 
After label-persistence smoothing, these fragmented high-score regions are connected into a more coherent span, producing predictions that better reflect the span-level nature of hallucination annotations. 
At the same time, this example also illustrates a trade-off introduced by smoothing: when hallucinated evidence appears in several nearby but distinct regions, label-persistence smoothing may merge them into a broader continuous span. 
Thus, label-persistence smoothing improves span consistency, but may reduce boundary precision in cases where hallucinated content is interleaved with faithful tokens.

The same example also illustrates the role of the contextual residual \(c\). 
When \(c\) is removed, the model tends to assign high scores more broadly in regions where the current token is influenced by preceding generated context. 
This behavior is consistent with the motivation for the residual feature: \(\Delta h\) alone captures reference-induced changes at the current token, but it does not explicitly account for reference influence that has already been expressed through earlier answer tokens. 
By incorporating \(c\), CORTEX can distinguish direct token-level reference sensitivity from deviations relative to the preceding context, leading to more controlled localization.

Figure~\ref{fig:appendix_ablation_advice} shows a different type of error case in which the answer includes additional advice that is not supported by the provided references. 
The advice is informative and may be reasonable from the model's parametric knowledge, but it is not grounded in the supplied passages. 
CORTEX assigns high hallucination scores to this region because its objective is to detect whether the answer is supported by the references, not whether the content is generally plausible or useful.

This case highlights an important ambiguity in reference-grounded hallucination detection. 
In controlled RAG applications, such as enterprise customer support, advice that is not grounded in references can be undesirable or risky even when it is factually plausible, because the system is expected to answer only from those documents.
In such settings, flagging unsupported advice is a desirable behavior. 
In contrast, in open-ended assistant scenarios, suppressing all useful but reference-unsupported content may overly restrict the capabilities of the LLM. 
Therefore, the appropriate treatment of such content depends on the application: CORTEX should be interpreted as a detector of reference support, rather than a general judge of factual correctness or utility.

This example also clarifies the scope of CORTEX. 
Because the method compares internal representations under inputs with and without references, it is sensitive to whether a token is grounded in the provided evidence. 
Consequently, content generated from parametric knowledge alone can receive high hallucination scores if it is not supported by the references. 
This behavior is aligned with the intended design of CORTEX for reference-grounded generation, but it should be considered when applying the method to settings where answers are allowed to go beyond the retrieved references.

\end{document}